\documentclass[lettersize,journal]{IEEEtran}
\usepackage{amsmath,amsfonts}
\usepackage{algorithmic}
\usepackage{algorithm}
\usepackage{array}
\usepackage[caption=false,font=normalsize,labelfont=sf,textfont=sf]{subfig}
\usepackage{textcomp}
\usepackage{stfloats}
\usepackage{url}
\usepackage{verbatim}
\usepackage{graphicx}
\usepackage{cite}
\usepackage{multirow}
\usepackage{hyperref}
\usepackage{multicol}
\usepackage{color,colortbl}
\usepackage[justification=centering]{caption}
\begin{document}

\title{Building Extraction from Remote Sensing Images via an Uncertainty-Aware Network}

\author{Wei He,~\IEEEmembership{Senior Member,~IEEE}, Jiepan Li, Weinan Cao, Liangpei Zhang,~\IEEEmembership{Fellow,~IEEE}, Hongyan Zhang, ~\IEEEmembership{Senior Member,~IEEE}
\thanks{W. He, J. Li, W. Cao, and L. Zhang are with the State Key Laboratory of Information Engineering in Surveying, Mapping and Remote Sensing, Wuhan University, Wuhan 430072, China (\{weihe1990, jiepanli, jackzhang, zlp62\}@whu.edu.cn).}
\thanks{H. Zhang is with the School of Computer Science, China University of Geosciences, Wuhan, 430074, PR China (zhanghongyan@whu.edu.cn).}
}



\maketitle

\begin{abstract}
Building extraction aims to segment building pixels from remote sensing images  and plays an essential role in many applications, such as city planning and urban dynamic monitoring. Over the past few years, deep learning methods with encoder–decoder architectures have achieved remarkable performance due to their powerful feature representation capability. Nevertheless, due to the varying scales and styles of buildings, conventional deep learning models always suffer from uncertain predictions and cannot accurately distinguish the complete footprints of the building from the complex distribution of ground objects, leading to a large degree of omission and commission. In this paper, we realize the importance of uncertain prediction and propose a novel and straightforward Uncertainty-Aware Network (UANet) to alleviate this problem. Specifically, we first apply a general encoder–decoder network to obtain a building extraction map with relatively high uncertainty. Second, in order to aggregate the useful information in the highest-level features, we design a Prior Information Guide Module to guide the highest-level features in learning the prior information from the conventional extraction map. Third, based on the uncertain extraction map, we introduce an Uncertainty Rank Algorithm to measure the uncertainty level of each pixel belonging to the foreground and the background. We further combine this algorithm with the proposed Uncertainty-Aware Fusion Module to facilitate level-by-level feature refinement and obtain the final refined extraction map with low uncertainty. To verify the performance of our proposed UANet, we conduct extensive experiments on three public building datasets, including the WHU building dataset, the Massachusetts building dataset, and the Inria aerial image dataset. Results demonstrate that the proposed UANet outperforms other state-of-the-art algorithms by a large margin.  The source code of the proposed UANet is available at \href{https://github.com/Henryjiepanli/Uncertainty-aware-Network}{https://github.com/Henryjiepanli/Uncertainty-aware-Network}.

\end{abstract}


\section{Introduction}
\IEEEPARstart{B}{uilding} extraction aims to distinguish building footprints from high-resolution remote sensing (RS) images, and has made remarkable progress in the past few decades. Owing to its potential applications, building extraction has also been extended to various fields, such as city planning \cite{urban_planning}, urban dynamic monitoring \cite{urban_monitor}, and disaster detection \cite{disaster_detection}.

\begin{figure}[t]
\vspace{-2.0em}
\centering
\subfloat[Image]{
\includegraphics[width = 2.0cm,height=6cm]{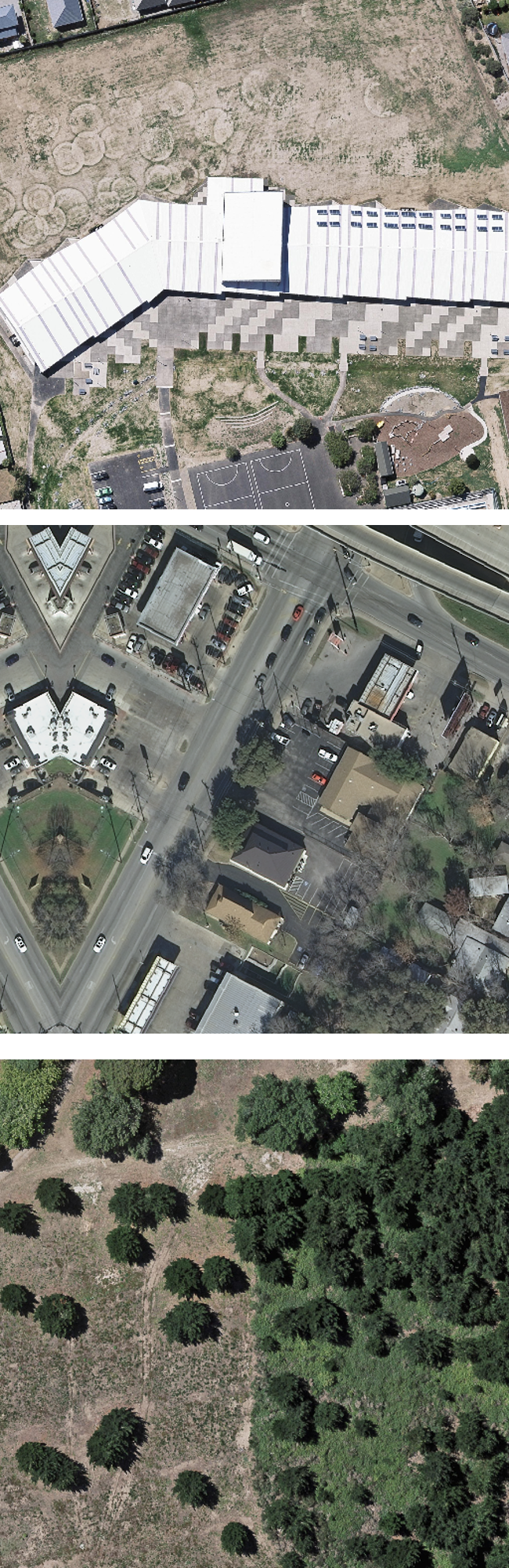}%
\label{fig_1_IMG}}
\hfil
\subfloat[GT]{
\includegraphics[width = 2.0cm,height=6cm]{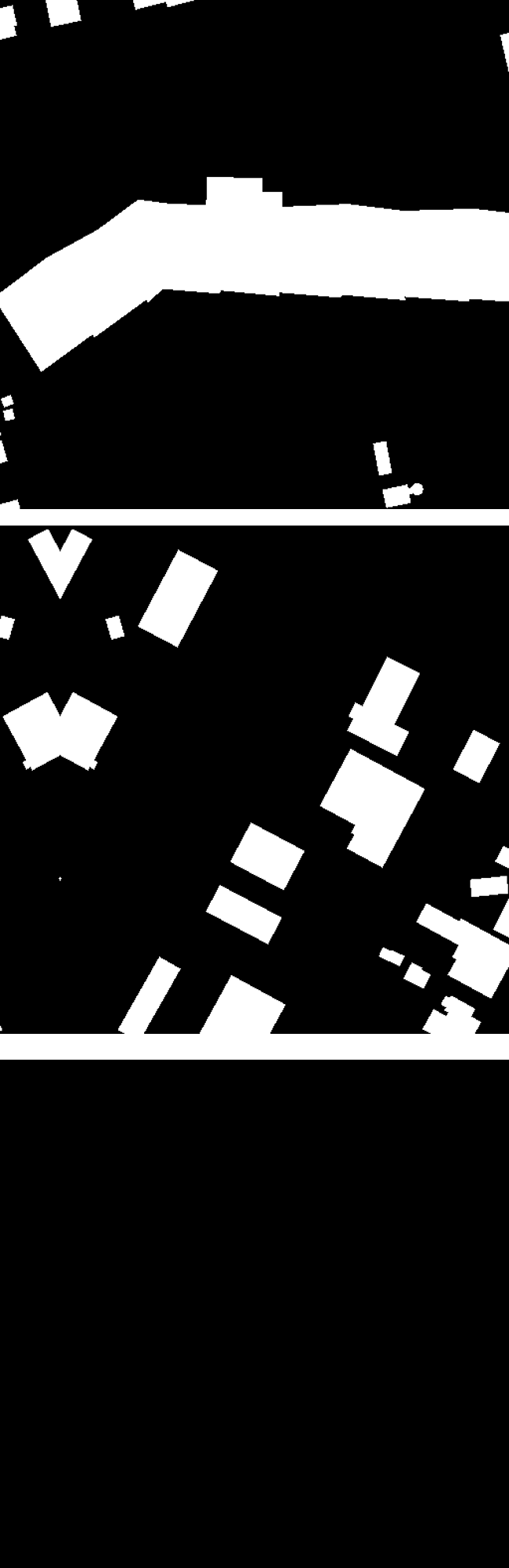}%
\label{fig_1_GTl}}
\subfloat[SOTA]{
\includegraphics[width = 2.0cm,height=6cm]{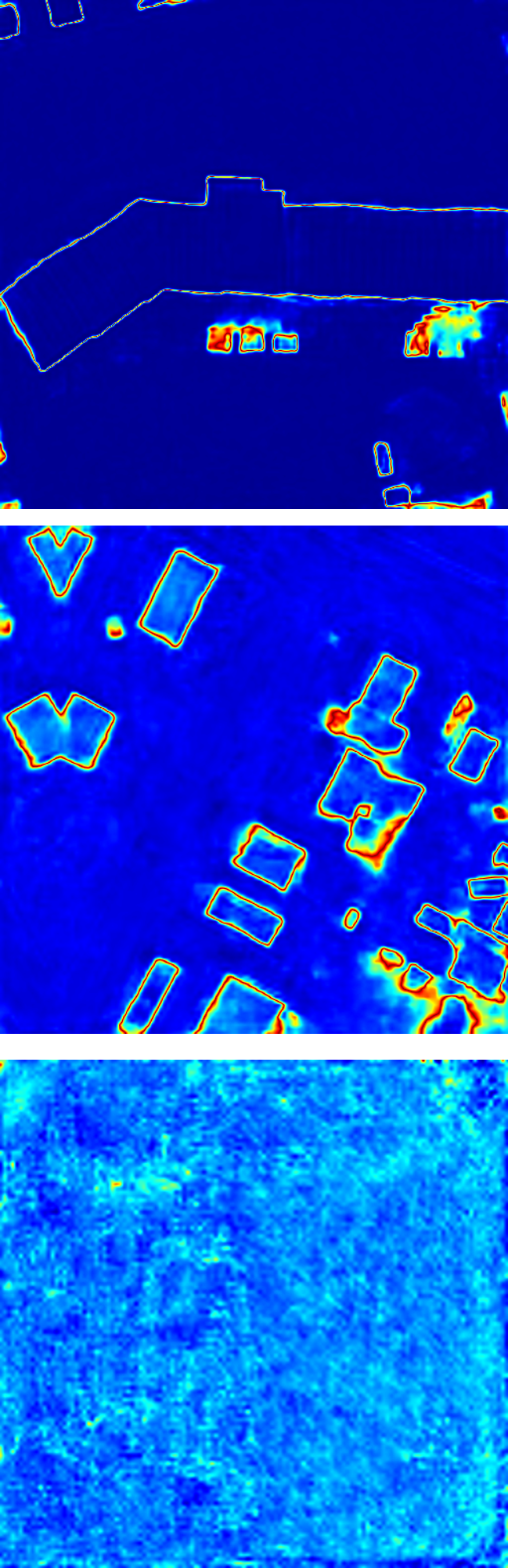}%
\label{fig_1_SOTA}}
\hfil
\subfloat[Ours]{
\includegraphics[height=6cm]{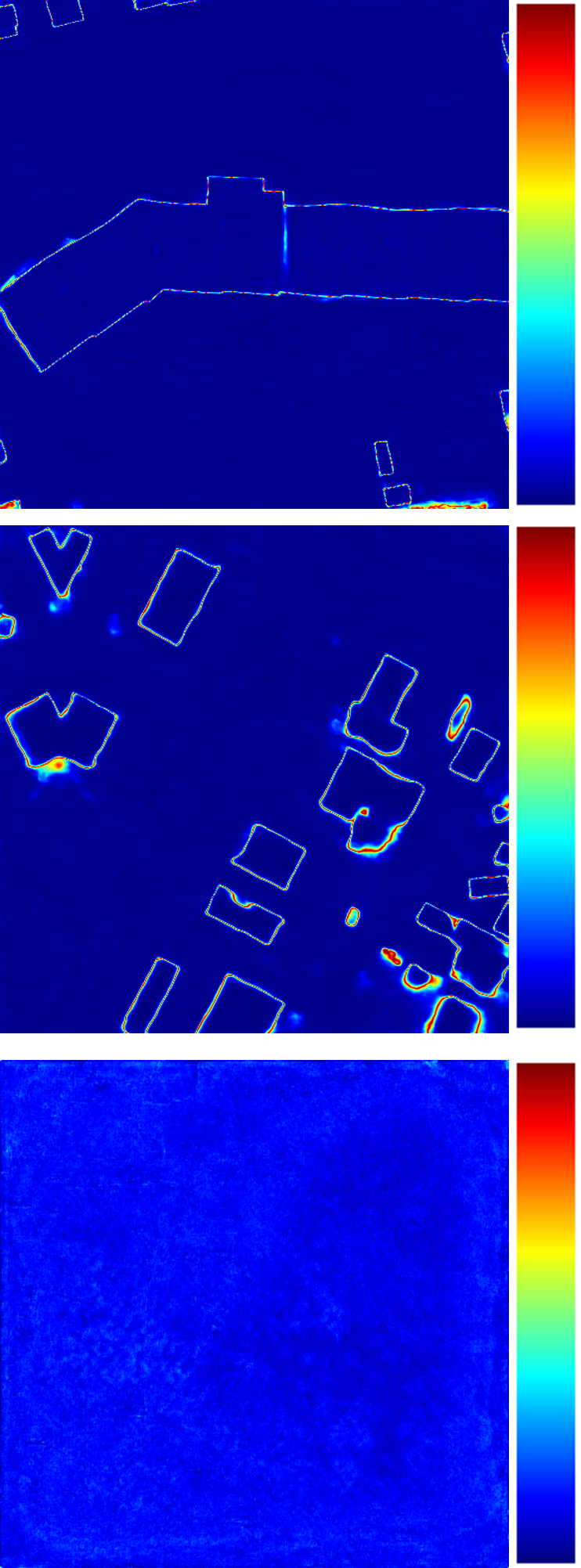}%
\label{fig_1_ours}}

\caption{Uncertainty visualizations between our proposed UANet and the state-of-the-art (SOTA) method for building extraction (BuildFormer \cite{BuildFormer}). (c) and (d) are achieved by the operation $0.5-|0.5-\star|$, with $\star$ representing the output of the $Sigmoid$ function.}
\vspace{-2.em}
\label{fig_1}
\end{figure}
\par
Up to date, numerous studies have made significant contributions to the extraction of buildings from high-resolution remote sensing (RS) images (\cite{DSNet,BuildFormer,MA_FCN}). Compared to middle/low resolution RS images, the higher-resolution RS images provide more detailed information about ground objects, while also increasing intra-class variances and decreasing inter-class variances, posing various challenges to accurately extract building footprints \cite{variance_inter_intra}.
To overcome the aforementioned challenges, research on building extraction has undergone a long-term development. In the early stage, a major effort was devoted to the design of more distinctive features. For example, \cite{color} utilized multiple colors and color-invariant spaces to select the representative corners and chose some corner candidates to generate the rooftop outline. Based on information about entropy and color, \cite{texture} introduced texture information to differentiate between buildings and trees. Moreover, \cite{contour} firstly took advantage of the contour driven by edge-flow to extract the building boundary, and then segmented the compositional polygons of the building roof by Joint Systems Engineering Group (JSEG). Nevertheless, due to the limited robustness and representativeness, the aforementioned hand-crafted features cannot handle the complex correlation between the buildings and the background.

\par
In the past few years, deep learning algorithms have been successfully applied to RS building extraction and have become the mainstream technical tools. Initially, in order to adopt deep learning algorithms into building extraction research, some simple networks were proposed based on patch-based annotation. \cite{MOE} designed a neural network consisting of three convolutional layers and two fully connected layers to achieve the automatic extraction of buildings. \cite{Simultaneous} designed a patch-based convolutional neural network (CNN) architecture that replaced fully connected layers with global average pooling (GAP) to improve extraction performance. However, the patch-based classification method has two unavoidable drawbacks \cite{luo2021deep}, namely, a huge computational burden and limited long-distance information exchange. As a result, these methods cannot fully exploit contextual information in high-resolution RS images, making it difficult to completely and accurately extract buildings from complex backgrounds.

\begin{figure*}[t]
\centering
\includegraphics[width=1.0\linewidth]{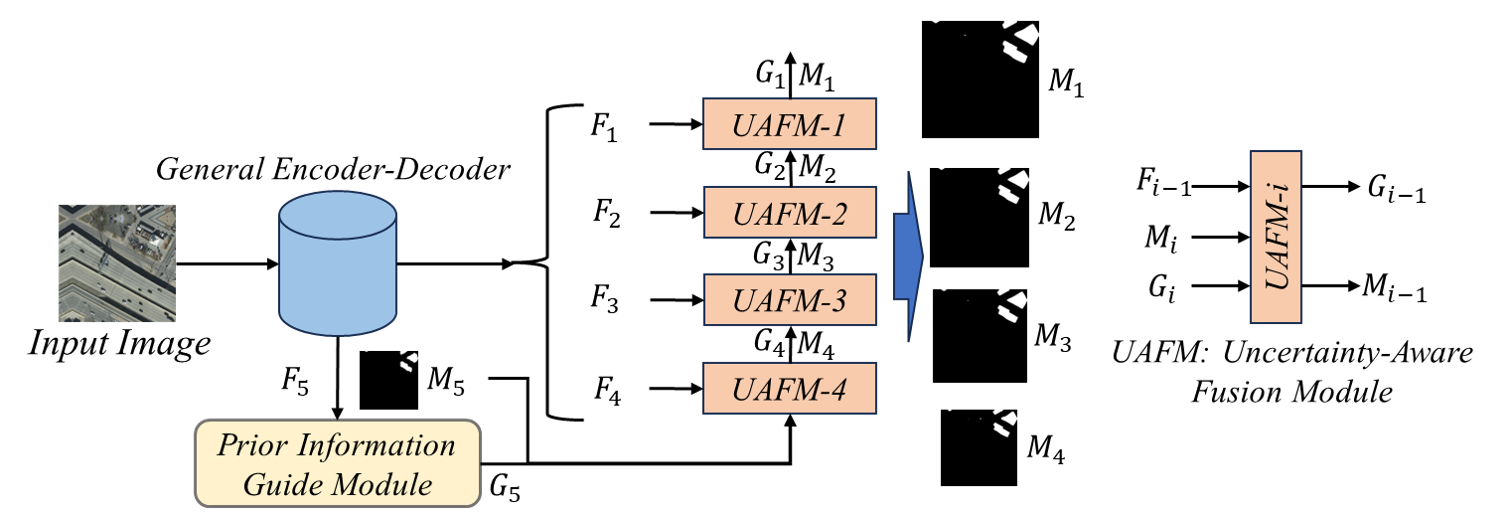}
\caption{The structure of the Uncertainty-Aware Network, which is composed of a general encoder-decoder, a prior information guide module (PIGM), and an uncertainty-aware fusion module (UAFM). }
\vspace{-1.0em}
\label{fig_2}
\end{figure*}
\par

Fully Convolutional Network (FCN) \cite{FCN} is a landmark pixel-based segmentation method that provides an encoder-decoder architecture, which has become a paradigm. In detail, the encoders process the input image to generate multi-level features, and the decoders adopt various strategies to output the semantic results. Currently, typical backbone networks, such as VGG \cite{VGG}, ResNet \cite{ResNet}, ResNext \cite{ResNext}, Res2Net \cite{Res2Net}, and even some networks based on transformers \cite{SwinTransformer, PVT}, are selected as encoders. After obtaining hierarchical features from the encoder, a sequence of decoder structures is proposed. For designing the decoders, the general strategy is to take advantage of multi-level encoded features from the aspects of modeling multi-scale contextual information \cite{Deeplabv3+,RFB,small_object_segmentation,ASF-Net,MSNet,fang2022context}, mining long-range dependency information \cite{Self-attention-building,Non-local,DANet,DSNet,SAB_Net,song2023denoised}, or feature refinement \cite{CBRNet,FPCRF,DAD}.

\par
Regarding the decoding strategies for modeling multi-scale contextual information, two typical plug-and-play modules, namely, Atrous Spatial Pyramid Pooling (ASPP) \cite{Deeplabv3+} and Receptive Field Block (RFB) \cite{RFB}, have been proposed. Furthermore, \cite{small_object_segmentation} enhanced the extraction of local features with a reasonable stacking of small-dilation-rate dilation convolutions, thereby effectively reducing the cases of ambiguous results for small-sized building segmentation. \cite{ASF-Net} proposed a novel Adaptive Screening Feature Network to teach the network to adjust the receptive field and adaptively enhance useful feature information. Moreover, \cite{MSNet} utilized a graph-based scale-aware structure to model and reason the interactions between different scale features.

\par
Regarding the decoding strategies for mining long-range dependency information, there have been many great works on the design of both encoders and decoders. Given that CNN is limited to local connections, some researchers have replaced CNNs with transformers in the design of encoders. The current transformer networks, such as Swin Transformer \cite{SwinTransformer}, Pyramid Vision Transformer \cite{PVT}, and so on, have all proved their strength in capturing long-distance information. Additionally, some work has introduced unique modules to establish long-distance contextual information in the decoders. For example, an Asymmetric Pyramid Non-Local Block \cite{APNB} was introduced by \cite{Non-local-building} to extract contextual global information. \cite{Self-attention-building} combined an ASPP \cite{Deeplabv3+} and a Non-Local Block \cite{Non-local} to propose a pyramidal self-attentive module for convenient embedding in the network. \cite{DSNet} took advantage of a local-global dual-stream network to adaptively capture local and long-range information for accurate building extraction.

\par
Regarding the decoding strategies for feature refinement, researchers are suggested to model the long distances and accurately locate spatial locations, which is often overlooked by CNN due to the spatial transformation invariance. Therefore, some works have utilized boundaries and contours to refine the final segmentation. \cite{FPCRF} proposed the Feature-Pairwise Conditional Random Field based on the Graph Convolutional Network (GCN) \cite{GNN}, which is a conditional random field for pairs of potential pixels with local constraints, incorporating the feature maps extracted by the CNN. \cite{gatedGCN_boundary} analyzed the conflict between deep CNN downsampling operation and accurate boundary segmentation, and introduced a Gated GCN into the CNN structure to generate clear boundaries and high fine-grained pixel-level classification results. Furthermore, \cite{CBRNet} designed a boundary refinement module (BR) to progressively refine the prediction of the building by perceiving and refining the edges of the building. Although taking the boundary information into account seems to be an appropriate way to refine the details of the segmentation, the richness of the boundary samples can also be an important factor that should not be ignored to limit the performance exploration of such methods.

\par
In summary, great progress has been made for high resolution RS building extraction using deep-learning-based methods. However, due to the complex distribution of ground objects in RS images and diverse appearances of buildings, current decoding strategies will inevitably produce misunderstanding of the building, resulting in uncertain prediction, which is clearly reflected in Fig. \ref{fig_1_SOTA}.
As analyzed in \cite{What_uncertainties}, the reason why current decoding strategies fail in some difficult cases is that they lack enough attention to hard-to-segment samples. Especially in RS images, some buildings are not salient enough and do not appear frequently, which will result in uncertainty of the model.
Therefore, solving the uncertain prediction is the key to further improving the performance of building extraction model.
In fact, uncertainty-aware learning has been studied in the general segmentation \cite{Phiseg,Assessing_reliability,UGTR} and detection \cite{kraus2019uncertainty} areas. 
At the beginning, the uncertainty analysis is always tight to complex networks (Bayesian deep learning \cite{What_uncertainties,bayesian_uncertainty} $,etc.,$) with a huge computational cost.
Subsequently, the general frameworks (Probabilistic Representational Network \cite{UDNet,UGTR}, Generative Adversarial Network (GAN) $,etc.$) are designed to improve the prediction certainty.
However, when dealing with a building extraction task,
these models/frameworks may fail to explore the characteristics of RS and result in unsatisfied results. 

\par
In this paper, we realize the importance of building uncertainty prediction, and propose the Uncertainty-Aware Network (UANet).
The proposed UANet can automatically rank the background uncertainty and building uncertainty of RS, and progressively guide the attention to these uncertain pixels during the interaction of features.
In detail, the proposed UANet first adopts a conventional encoder-decoder structure to output multi-level features and a relatively uncertain extraction map. On the basis of these results, we attempt to further solve the uncertainty problem and divide the following process into two key parts.
At the beginning, we put forward a prior information guide module (PIGM) via a novel cross-attention mechanism to realize the enhancement of both spatial and channel aspects. Then, we propose the uncertainty-aware fusion module (UAFM) and innovatively invent an uncertainty rank algorithm (URA) to realize the elimination of uncertainty as much as possible. As shown in Figs. \ref{fig_1_SOTA} and \ref{fig_1_ours}, compared with BuildFormer, our UANet shows less uncertainty particularly around the edges. The main contributions of this study are as follows.

\begin{enumerate}[]
\item We introduce the uncertainty concept to building extraction and propose the UANet that can maintain high certainty faced with diverse scales, complex backgrounds, and various building appearances, $etc$. 
\item We put forward a novel feature refinement way named PIGM from both spatial and channel aspects.
\item We propose the UAFM and the URA to relieve the uncertainty condition and achieve a refined extraction map with low uncertainty.
\end{enumerate}
\par
The rest of this paper is organized as follows. In Section 2, we analyze and introduce the structure and components of our UANet. The experiments and results analysis are presented in Section 3, the ablation study of our proposed modules is given in Section 4, and the conclusions are outlined in Section 5.

\section{Methodology}
\subsection{Overview}
Aiming to eliminate the uncertainty of the final extraction map as much as possible, we propose the Uncertainty-Aware Network (UANet). As shown in Fig.\ref{fig_2}, we first adopt a general encoder-decoder network to get a relatively uncertain extraction map. Regarding the general encoder-decoder part, we adopt VGG-16 \cite{VGG} as the encoder backbone to extract multi-level features from the input image, introduce a multi-branch dilation convolution blocks to enhance the encoded features ($E_{i}, \left\{i=2,3,4,5 \right \}$), and use a typical cross-fusion strategy (Feature Pyramid Network (FPN \cite{FPN})) to obtain a relatively uncertain extraction map $M_{5}$. Based on the output features ($F_{i}, \left\{i=1,2,3,4,5 \right \}$) and uncertain extraction map $M_{5}$, our UANet acts as a decoder strategy to deal with the general building extraction challenges and output a refined extraction map with low uncertainty.
In detail, we first put forward a prior information guide module (PIGM) to take advantage of the prior information of the obtained extraction map to enhance the highest-level feature. Subsequently, the uncertainty-aware fusion module (UAFM) is utilized progressively to eliminate the uncertainty of features from high level to low level. Finally, UANet outputs the final refined extraction map with lower uncertainty.

\subsection{Prior Information Guide Module}

\begin{figure*}[t]
\centering
\includegraphics[width=1.0\linewidth]{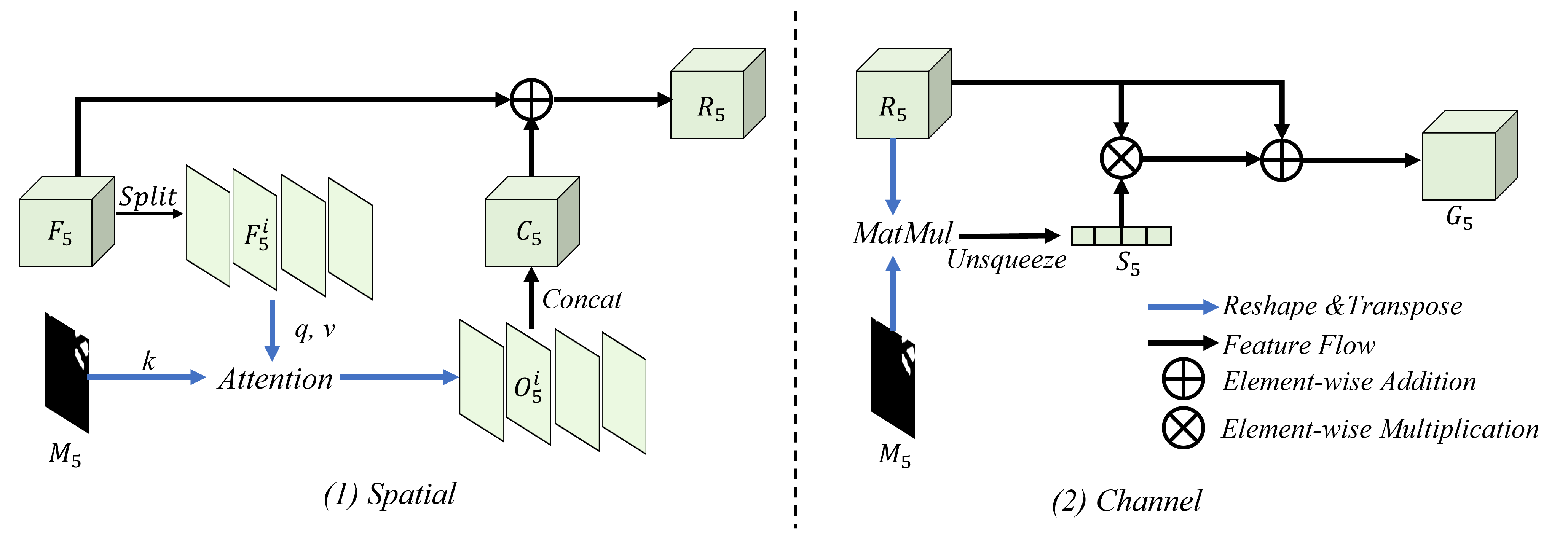}
\caption{\centering{The structure of the Prior Information Guide Module (PIGM).}}
\vspace{-1.0em}
\label{fig_4}
\end{figure*}

In fact, the process to achieve the relatively uncertain extraction map $M_5$ is a general decoding strategy, which cannot solve the current uncertainty problem. However, we believe that the information provided by $M_5$ is still very valuable. Therefore, to achieve a more accurate and less uncertain prediction, we try to consider the extraction map $M_5$ as prior knowledge and take advantage of it to realize the enhancement of the features. As mentioned before, the highest-level feature with the largest dimension lacks spatial information due to the smallest resolution. Based on this consideration, we propose the Prior Information Guide Module (PIGM) to guide the highest-level feature to realize refinement from both spatial and channel aspects. As shown in Fig.~\ref{fig_4}, we first utilize $M_5$ to guide the highest-level feature to learn the corresponding spatial relationships. Subsequently, we continue to use $M_5$ to model the channel dependence of the enhanced feature.

\par
In detail, the inputs of PIGM are $F_{5} \in  {\mathbb{R}}^ {C\times H\times W}$ and $M_{5} \in  {\mathbb{R}}^ {1\times H\times W}$. At the beginning, we split the input feature $F_{5}$ from the channel dimension and get $C$ feature maps $F_{5}^{i} \in {\mathbb{R}}^ {1\times H\times W} $:
\begin{equation}
\label{equ1} 
    F_{5}^{i} = Split(F_{5}), i=1,2,...,C.
\end{equation}

\par
On the one hand, we reshape $F_{5}^{i}$ to compress its dimension and obtain $V_{5}^{i} \in  {\mathbb{R}}^ {C\times N}$ ($N = H \times W$). On the other hand, we reshape and transpose $F_{5}^{i}$ to get $Q_{5}^{i} \in  {\mathbb{R}}^ {N\times C}$:
\begin{equation}
\label{equ2} 
\begin{split}
    &V_{5}^{i} = Reshape(F_{5}^{i}), i=1,2,...,n,\\
    &Q_{5}^{i} = Transpose(Reshape(F_{5}^{i})), i=1,2,...,n.
\end{split}
\end{equation}

\par
Subsequently, we need to guide the input feature $F_{5}$ to learn the spatial information by exploring the prior map $M_{5}$, so we reshape $M_{5}$ to $K_{5} \in  {\mathbb{R}}^ {N\times C}$:

\begin{equation}
\centering
\label{equ3} 
    K_{5} = Reshape(M_{5}).
\end{equation}
\par
Then, we perform the cross-attention, which conducts the matrix multiplication between $Q_{5}^{i}$ and $K_{5}$ via Softmax function to obtain $T_{5}^{i} \in  {\mathbb{R}}^ {N\times N}$ that represents the relationship between each channel of $F_{5}$ and the prior map $P_5$:
\begin{equation}
\centering
\label{equ4} 
\begin{split}
    &T_{5}^{i} = Softmax(Q_{5}^{i}\otimes K_{5}), i=1,2,...,n.
\end{split}
\end{equation}
\par
We take use of the relation map $T_{5}^{i}$ to multiply $V_{5}^{i}$ and achieve the enhanced feature map $O_{5}^{i} \in {\mathbb{R}}^ {1\times H\times W}$. Meanwhile, all the enhanced feature maps from $O_{5}^{1}$ to $O_{5}^{C}$ are concatenated to formulate $C_{5}$. At last, we can obtain the spatial enhanced feature $R_{5} \in {\mathbb{R}}^ {C\times H\times W}$ with a residual structure:
\begin{equation}
\centering
\label{equ5} 
\begin{split}
    &O_{5}^{i} = V_{5}^{i}\otimes T_{5}^{i}, i=1,2,...,C, \\
    &C_{5} = Concat(O_{5}^{0},O_{5}^{i},...,O_{5}^{C}), \\
    &R_{5} = \alpha \times C_{5} + F_{5},
\end{split}
\end{equation}
where $\alpha$ is a learnable parameter.
\par
From another perspective, we still attempt to utilize $M_{5}$ to achieve the enhancement of $R_5$. As shown in Fig.~\ref{fig_4}, we reshape and transpose $M_{5}$ to obtain $Q_{5}^{'} \in  {\mathbb{R}}^ {N\times 1}$:
\begin{equation}
\centering
\label{equ6} 
\begin{split}
    &Q_{5}^{'} = Transpose(Reshape(M_{5})).
\end{split}
\end{equation}
At the same time, we reshape the spatial enhanced feature $R_{5}$ as $K_{5}^{'} \in  {\mathbb{R}}^ {C\times N}$, and use the matrix multiplication among $Q_{5}^{'}$ and $K_{5}^{'}$ and the Sigmoid function to get $S_{5} \in  {\mathbb{R}}^ {C\times 1}$. Afterward, by multiplying $S_{5}$ and $R_{5}$ associated with a residual structure, we can get the final feature $G_{5} \in  {\mathbb{R}}^ {C\times H \times W}$:

\begin{equation}
\centering
\label{pigm_FINAL} 
\begin{split}
    &K_{5}^{'} = Reshape(R_{5}), \\
    &S_{5} = Sigmoid(Q_{5}^{'}\otimes K_{5}^{'}), \\
    &G_{5} = \beta \times S_{5} \times R_{5} + R_{5}, 
\end{split}
\end{equation}
where $\beta$ is a learnable parameter.

\subsection{Uncertainty-Aware Fusion Module}
\label{sec:UAFM}
\begin{figure}[]
\centering
\includegraphics[width=1.0\linewidth]{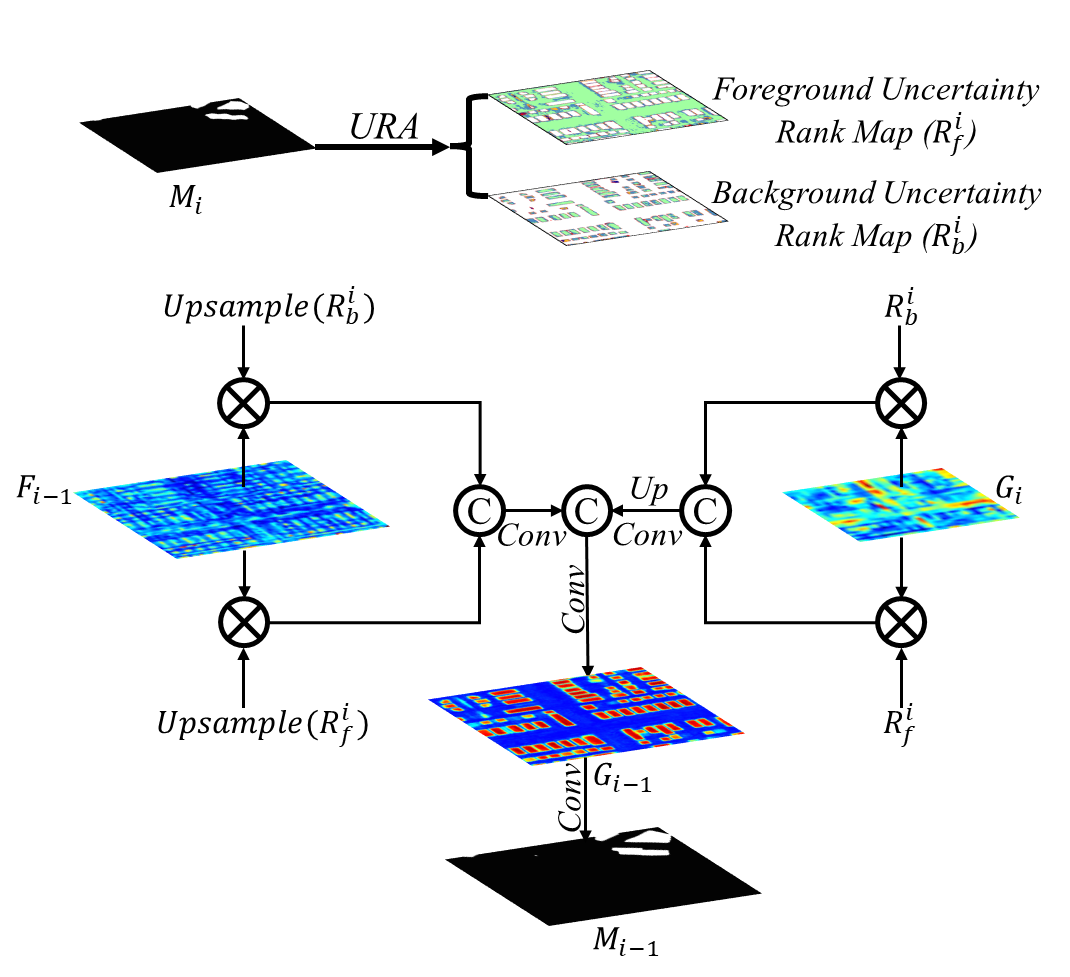}
\caption{The structure of the Uncertainty-Aware Fusion Module (UAFM).}
\vspace{-1.0em}
\label{fig_5}
\end{figure}

In the previous stages, we successively acquire the uncertain extraction map $M_{5}$ and the enhanced feature $G_{5}$. However, the uncertainty caused by the intricate backgrounds and various scales still remains. Therefore, we present the uncertainty-aware fusion module (UAFM) to tackle the high uncertainty issue, as illustrated in Fig.~\ref{fig_5}.
\par
As we all know, all the deep learning approaches output the extraction results by using the $Softmax$ function to allocate the corresponding probability for each pixel, which can be directly used to reflect the uncertainty of the model in its predictions. As mentioned before, in RS images, some buildings are not salient enough and do not appear frequently, which will result in the uncertainty of model. To overcome such a uncertainty problem, we directly use the $Sigmoid$ function to get the corresponding probabilities of all pixels in the extraction map $M$ from spatial perspective, then we subtract all values of the probability map with $0.5$ to measure the uncertainty belonging to foreground ($U_{f}$) and meanwhile subtract $0.5$ with all values of the probability map to measure the uncertainty belonging to background ($U_{b}$),
\begin{equation}
\centering
\label{equ_u} 
\begin{split}
    &U_{f} = Sigmoid(M)-0.5, \\
    &U_{b} = 0.5 - Sigmoid(M).
\end{split}
\end{equation}
Subsequently, we rank the uncertainty of foreground and background into five levels using the Uncertainty Rank Algorithm (URA), that is, the range of $[-0.5,0)$ represents not in consideration (rank 0), the range of $[0, 0.1)$ indicates the highest uncertainty (rank 5), the range of $[0.1, 0.2)$ represents the relatively high uncertainty (rank 4), the range of $[0.2, 0.3)$ represents the central uncertainty (rank 3), the range of $[0.3, 0.4)$ indicates moderately low uncertainty (rank 2), and the range of $[0.4, 0.5]$ denotes the lowest uncertainty (rank 1). We then assign corresponding uncertainty levels as weights to the pixels, with the principle of attaching higher weights to pixels with higher uncertainty, so as to pay more attention on uncertain areas. We denote URA as:
\begin{equation}
\label{equ10}
\mathbb{f}(i,j) = \left\{
\begin{array}{lr} \lfloor \frac {0.5 - U_{{i,j}}}{0.1}  \rfloor, U_{{i,j}} >= 0, & \\
0, U_{{i,j}} < 0, &
\end{array}
\right.
\end{equation}
where $U_{{i,j}}$ means the pixel in $i_{th}$ row and $j_{th}$ column of $U_{f}$ or $U_{b}$.

Therefore, after using URA to allocate the uncertainty level about the uncertainty maps of the foreground and the background, respectively, we can obtain the foreground uncertainty rank map ($R_{f}$) and the background uncertainty rank map ($R_{b}$).
\begin{equation}
\centering
\label{equ7-1} 
\begin{split}
    &R_{f} = URA(Sigmoid(M)-0.5), \\
    &R_{b} = URA(0.5 - Sigmoid(M)),
\end{split}
\end{equation}

\par
We take the fusion of the highest $G_{5}$ and $F_{4}$ for example to illustrate the whole fusion process. Specifically, the inputs of UAFM in this layer are the enhanced feature $G_5$, the $F_4$, and the uncertain extraction map $M_5$. Regarding the uncertainty-aware enhancement, on the one hand, we apply URA to $M_{5}$ so that we can get the corresponding foreground uncertainty rank map ($R_{f}^{5}$) and background uncertainty rank map ($R_{b}^{5}$). Then we directly use $G_{5}$ to multiply with them to highlight the uncertain pixels from both the foreground and background perspectives. Subsequently, we concatenate these two enhanced features and recover its original channel to get $G_{5}^{u}$ by a $1\times1$ convolution operation.
\begin{equation}
\centering
\begin{split}
\label{equ7-2} 
    &G_{5}^{u} = Conv_{1\times1}(Concat(R_{f}^{5}*G_{5},R_{b}^{5}*G_{5})),
\end{split}
\end{equation}
On the other hand, we use the nearest neighbor interpolation method to upsample $R_{f}^{5}$ and $R_{b}^{5}$ to the same size as $F_{4}$, and use the same operation to highlight $F_{4}$ as the enhancement of $G_{5}$, and we can get $F_{4}^{u}$. 
\begin{equation}
\centering
\label{equ7-3}
\begin{split}
    &F_{4}^{u} =Conv_{1\times1}(Concat(Up(R_{f}^{5})*F_{4},Up(R_{b}^{5})*F_{4})), \\
\end{split} 
\end{equation}
Finally, we upsample $G_{5}^{u}$ to match the size of $F_{4}^{u}$, concatenate them together, and use a $3\times3$ convolution operation to get the fused feature $G_{4}$, which can output the less uncertain extraction map $M_{4}$ by a $3\times3$ convolution operation.

\begin{equation}
\centering
\label{equ7-4}
\begin{split}
    & G_{4} = Conv_{3\times3}(Concat(F_{4}^{u},G_{5}^{u})), \\
    & M_{4} = Conv_{3\times3} (G_{4}),
\end{split} 
\end{equation}
\par
As shown in Fig. \ref{fig_5}, we employ the uncertainty-aware fusion module (UAFM) to fuse the features $G_i$ and $F_{i-1}$ layer-by-layer and decode the fused feature to output the corresponding certainty-improved map $M_{i-1}$.

With such a UAFM, we can utilize $M_4$ to fuse $G_{4}$ and $F_{3}$ and achieve output $M_3$, utilize $M_3$ to fuse $G_{3}$ and $F_{2}$ and achieve output $M_2$, and utilize $M_2$ to fuse $G_{2}$ and $F_{1}$ and output $M_1$, where $M_1$ can be viewed as the final refined extraction map with the lowest uncertainty.

\par
On the whole, we use the simple binary cross-entropy ($BCE$) loss function to supervise all the outputs, and the overall loss is :
\begin{equation}
\centering
Loss = \sum_{i=1}^{5} BCE(M_i, GT),
\end{equation}
where GT represents the ground truth.

\section{Experiments}
\subsection{Dataset}
To verify the superiority of our proposed UANet, we select three public building extraction datasets to conduct extensive experiments, including the WHU building dataset, the Massachusetts building dataset, and the Inria building dataset. The detailed information of the whole three datasets is described as follows:
\begin{enumerate}
\item  WHU building dataset \cite{whu_dataset} is composed of two types of images, $i.e.$, satellite images, and aerial images. In our experimental settings, we only conducted experiments on the aerial image dataset, which has $8,189$ image tiles ($4,736$ tiles for training, $1,036$ tiles for validation, and $2,416$ tiles for testing). The spatial resolution is just $0.3m$, and the whole aerial image dataset consists of $22,000$ buildings and covers a huge area of over $450 km^2$.
\item Massachusetts building dataset \cite{Massachusetts_building} owns 151 aerial images of the Boston area with spatial resolution $1m$. Composed of two types of scenes, $i.e.$, urban, and suburban, the Massachusetts building dataset covers almost $340 km^2$ areas, and all the image sizes are of $1500 \times 1500$ pixels. The official dataset contains a training set (137 images), a validation set (4 images), and a testing set (10 images). We adopt some data augmentation ways to expand the original training set to 411 images. For the training phase, we randomly crop the images and labels into 1024 × 1024 pixels as input. And for both the validating and testing phase, the images and labels are padded to the size of
1536 × 1536 pixels to ensure it is divisible by 32. It is worth mentioning that we ignore the padding parts when computing evaluation metrics.
\item Inria building dataset \cite{inria_dataset} contains 360 images collected from 5 cities (Austin, Chicago, Kitsap, Tyrol, and Vienna). Referring to the official suggestion, we select 1 to 5 tiles from each city for
validation and the rest for training. We first pad the original 5000 × 5000 images to 5120 × 5120 pixels and then crop them into 512 × 512 pixels image tiles. Second, we remove the images without buildings, the remaining 9737 and 1942 image tiles used for training and validation, respectively.

\end{enumerate}
\subsection{Evaluation Metrics}
\begin{table*}[]
\normalsize
\setlength\tabcolsep{3pt}
\caption{Performance comparison with baseline models on the test datasets. $\uparrow$ indicates the higher score the better and vice versa. The best score for each metric is marked in red. The second score for each metric is underlined.}
\label{tab:table_test}
\center
\begin{tabular}{c|c|cccc|cccc|cccc}
\hline
\multirow{2}{*}{Baseline}& \multirow{2}{*}{Year} & \multicolumn{4}{c|}{WHU (\%)}&\multicolumn{4}{c|}{Massachusetts (\%)} &\multicolumn{4}{c}{Inira (\%)}                                             
 \\ \cline{3-14} 
 & & $IoU\uparrow$ & $F1\uparrow$ & $Pre\uparrow$ & $Recall\uparrow$ 
 & $IoU\uparrow$ & $F1\uparrow$ & $Pre\uparrow$ & $Recall\uparrow$
 & $IoU\uparrow$ & $F1\uparrow$ & $Pre\uparrow$ & $Recall\uparrow$ \\ \hline
UNet \#& 2015 &
85.92 &92.39 &91.78 &93.01
&68.48	&81.47 &80.99	&81.96
&74.40	&85.32	&86.39	&84.28 \\
HRNet \#& 2019 &
85.64 &92.27 &91.69 &92.85
&69.39	&81.93	&81.49	&82.38
&75.03 &85.73 &86.56 &84.92\\ 
MA-FCN & 2019 &
90.70 &95.15 &95.20 &95.10
&73.80	&84.93	&87.07	&82.89
&79.67	&88.68	&89.82	&87.58 \\ 
DSNet \#& 2020 &
89.54 &94.48 &94.05 &94.91
&\underline{75.04}	&\underline{85.74}	&87.56	&83.99
&81.02	&89.52	&90.32	&88.73\\
CBRNet& 2021 &
\underline{91.40} &\underline{95.51} &\underline{95.31} &{\underline{95.70}}
&74.55	&85.42	&86.50	&84.36 
&81.10	&89.56	&89.93	&\underline{89.20}\\ 
MSNet & 2022 &
89.07 &93.96 &94.83 &93.12
&70.21	&79.33	&78.54	&80.14
& - &- &- &-\\ 
BOMSNet& 2022 &
90.15 &94.80 &95.14 &94.50
&74.71  &85.13	&86.64	&83.68
&78.18	&87.75	&87.93	&87.58\\ 
LCS & 2022 &
90.71 &95.12 &95.38 &94.86 
&- &- &- &-
&78.82	&88.15	&89.58	&86.77 \\
BuildFormer \#& 2022 &
90.73 &95.14 &{95.15} &95.14 
&75.03	&85.73	&86.69	&\underline{84.79}
&\underline{81.24}	&\underline{89.71}	&\underline{90.65}	&88.78
\\\
BCTNet & 2023 &
91.15 &95.37 &95.47 &95.27
&\underline{75.04} &\underline{85.74} &\underline{87.57} &83.99
&- & -&- &- \\
FD-Net &2023 &
91.14 & 95.36 & 95.27 &95.46
&74.54 &85.42 &87.95 &83.02
&- &- &- &- \\ \hline



\textbf{Ours-UANet} &  &
\color{red}\textbf{92.15} &{\color{red}\textbf{95.91}} &{\color{red}\textbf{95.96}} &\color{red}\textbf{95.86}

& \color{red}\textbf{{76.41}} &\color{red}\textbf{{86.63}} &\color{red}\textbf{87.94} &\color{red}\textbf{{85.35}}

& {\color{red}\textbf{83.08}} &{\color{red}\textbf{90.76}} &{\color{red}\textbf{92.04}} &{\color{red}\textbf{89.52}}
\\ \hline
\end{tabular}

{\textbf{\#} means that the results were obtained by ourselves. The codes of other compared methods are not released, we directly copy the results from the original papers. }
\vspace{-1.0em}
\end{table*}

\begin{figure*}
\centering
\subfloat[Image ]{
\includegraphics[width=2.3cm,height=7.5cm]{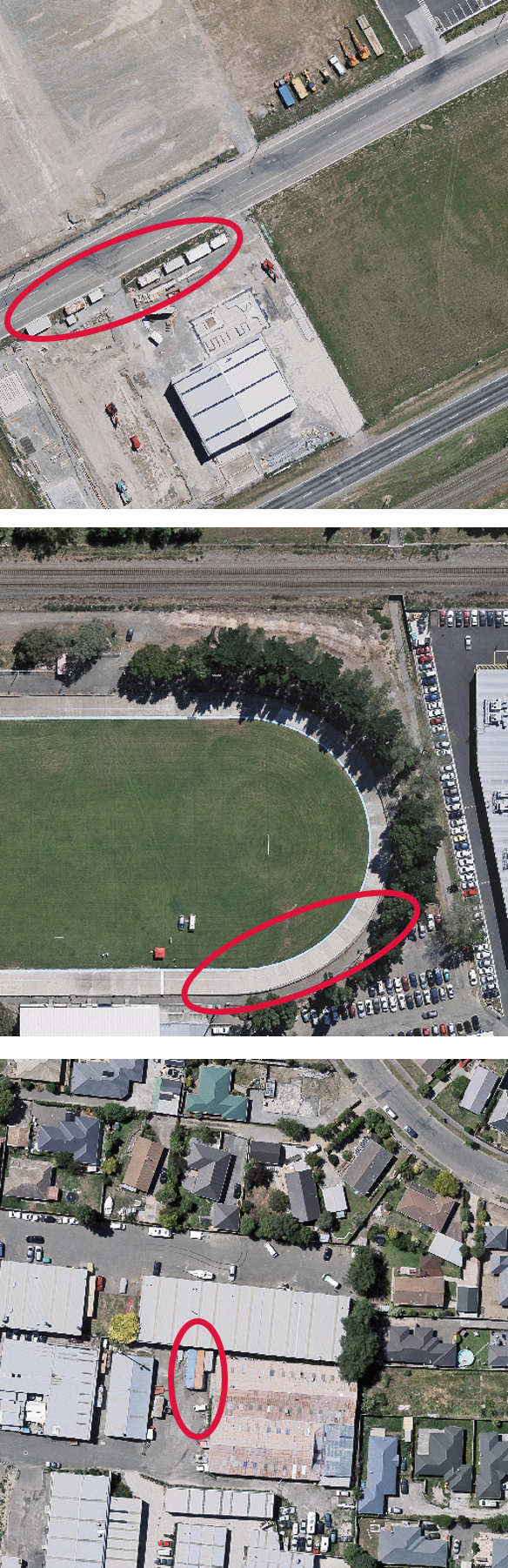}%
\label{fig_whu_tp_image}}
\hfil
\subfloat[GT]{
\includegraphics[width=2.3cm,height=7.5cm]{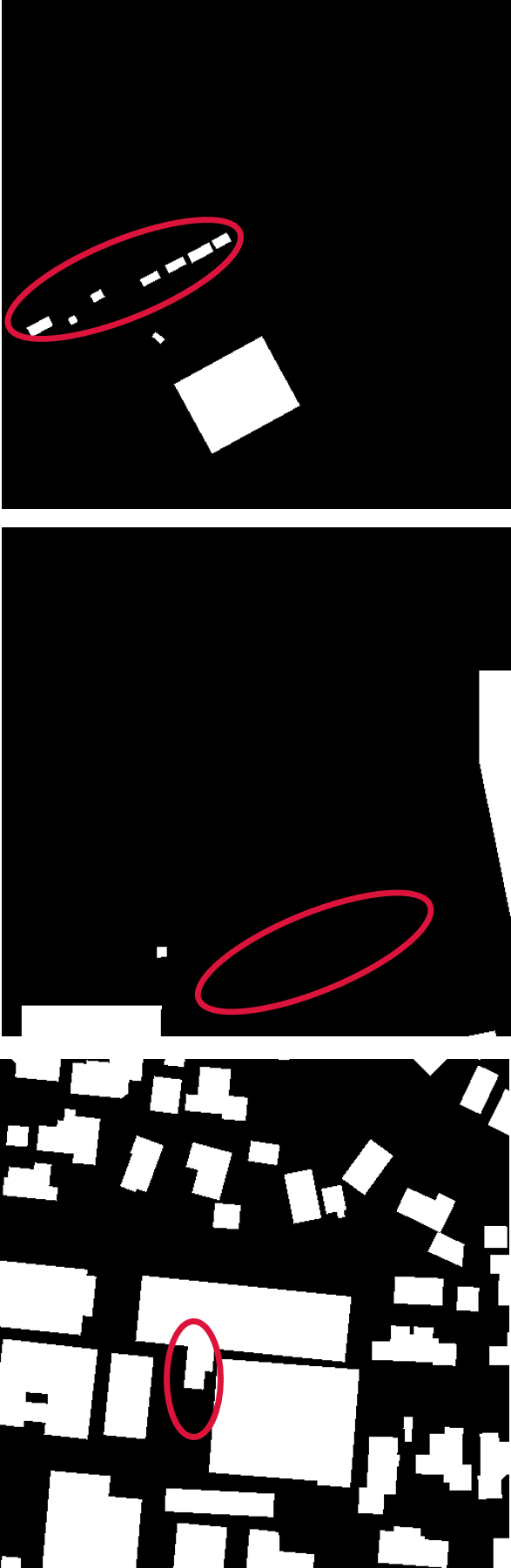}%
\label{fig_whu_tp_gt}}
\hfil
\subfloat[UNet]{
\includegraphics[width=2.3cm,height=7.5cm]{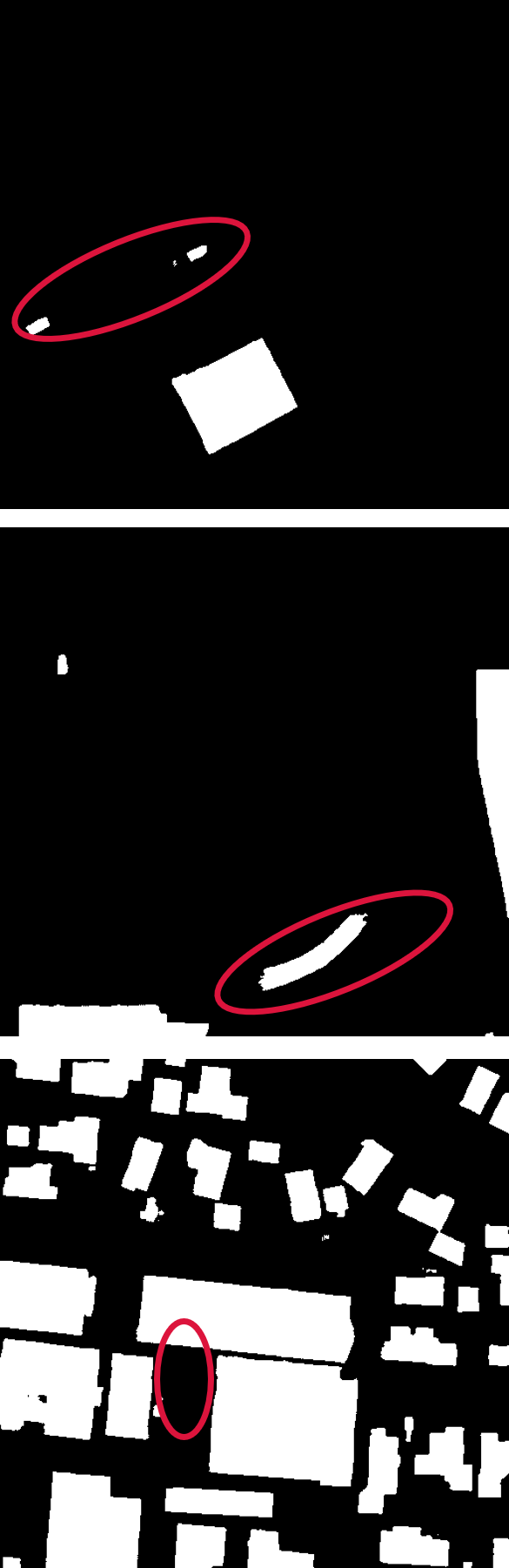}%
\label{fig_whu_tp_unet}}
\hfil
\subfloat[HRNet]{
\includegraphics[width=2.3cm,height=7.5cm]{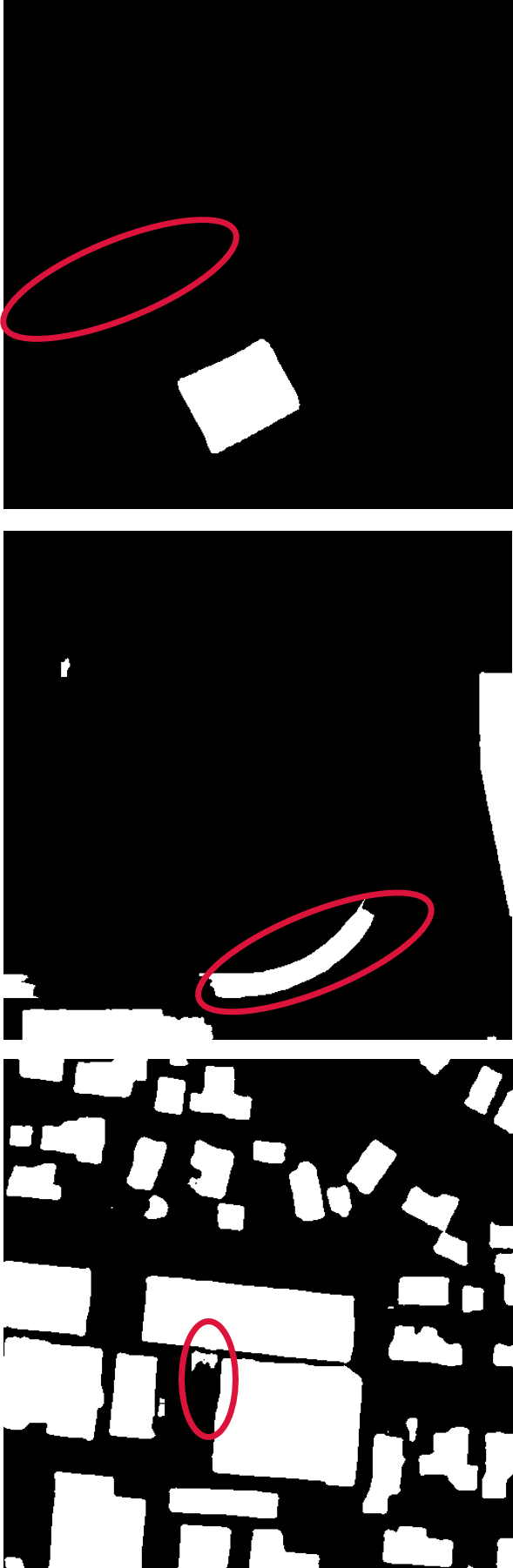}%
\label{fig_whu_tp_hrnet}}
\hfil
\subfloat[DSNet]{
\includegraphics[width=2.3cm,height=7.5cm]{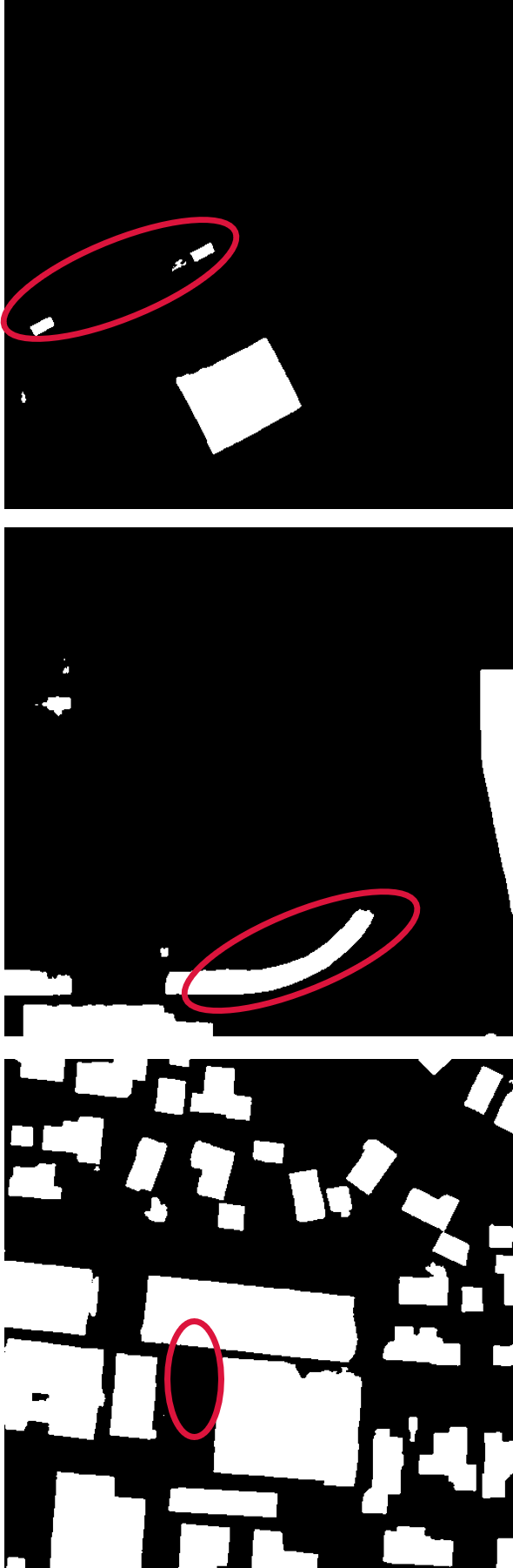}%
\label{fig_whu_tp_dsnet}}
\hfil
\subfloat[BuildFormer]{
\includegraphics[width=2.3cm,height=7.5cm]{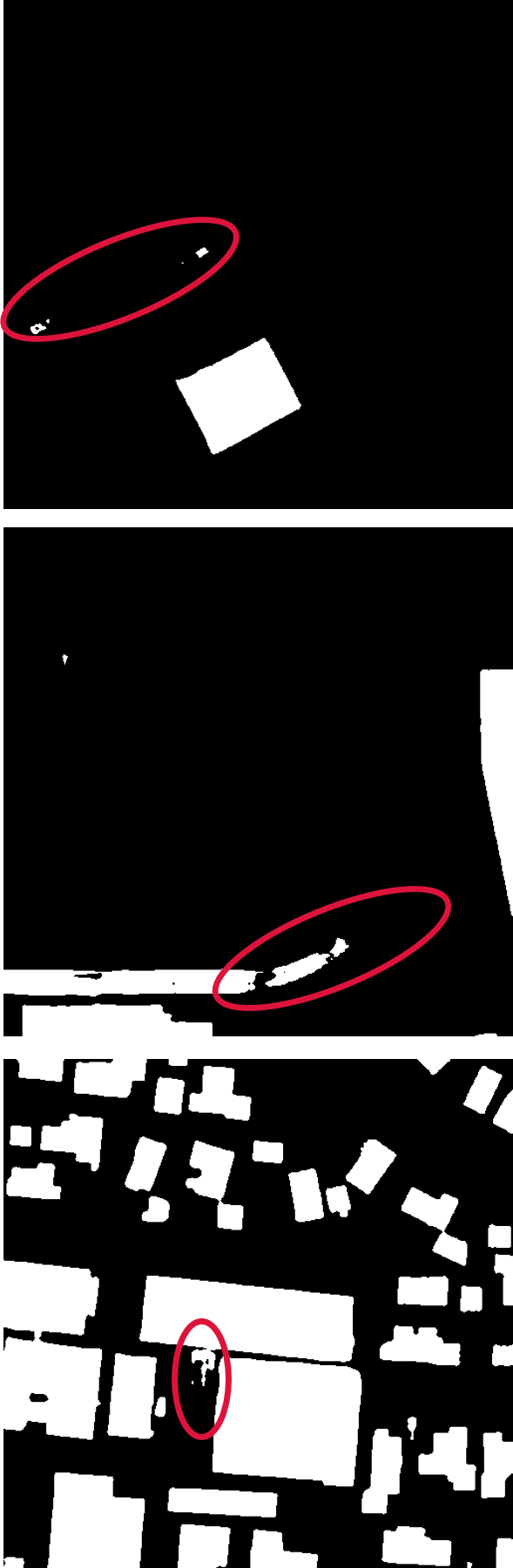}%
\label{fig_whu_tp_buildformer}}
\hfil
\subfloat[\textbf{Ours}]{
\includegraphics[width=2.3cm,height=7.5cm]
{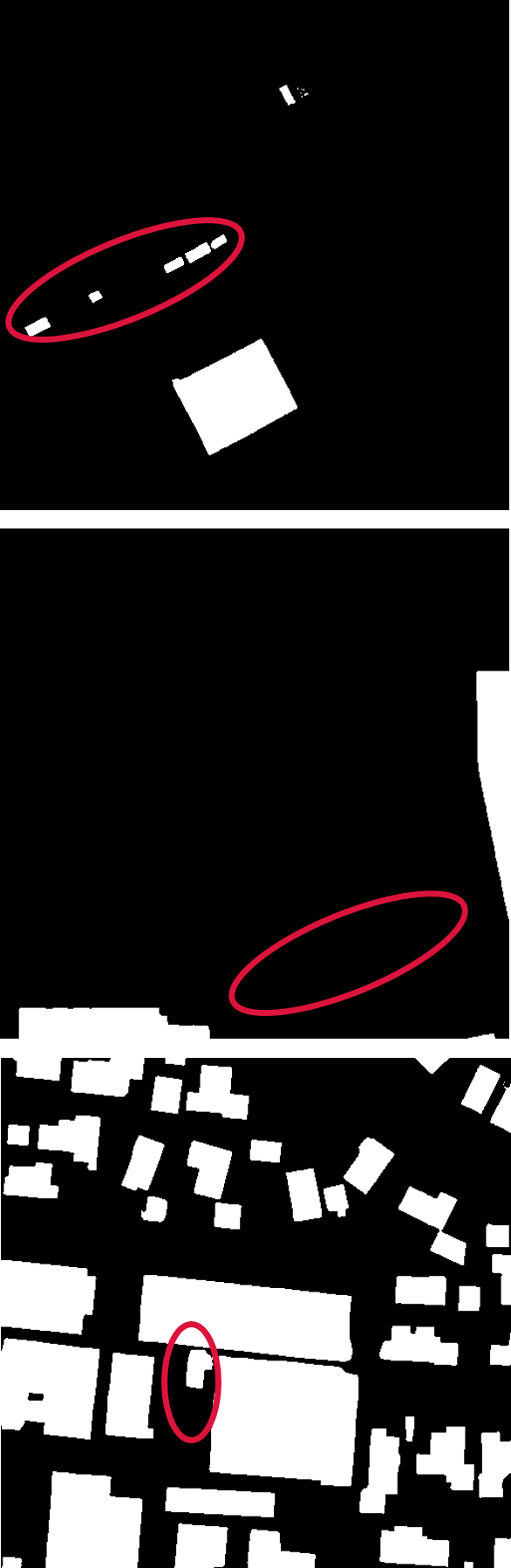}%
\label{fig_whu_tp_ours}}
\hfil
\caption{Visual Comparison on WHU building dataset.}
\label{fig_tp_whu}
\vspace{-0.5em}
\end{figure*}

To conduct a broad and comprehensive evaluation of our proposed model, we chose four metrics, $i.e.$, intersection over union ($IoU$), F1 score ($F1$), Precision, and Recall. At first, we use $TP$, $FP$, and $FN$ to represent the true positive, the false positive, and the false negative, respectively. Then, we give the definition of the four evaluation metrics as follows:

\begin{equation}
\centering
	\label{iou} 
	\begin{split}
 IoU = \frac{TP}{TP + FP + FN}
	\end{split}
\end{equation}
\vspace{-1.0em}
\begin{equation}
\centering
	\label{Pre} 
	\begin{split}
 Precision = \frac{TP}{TP + FP}
	\end{split}
\end{equation}
\vspace{-1.0em}
\begin{equation}
\centering
	\label{Recall} 
	\begin{split}
 Recall = \frac{TP}{TP + FN}
	\end{split}
\end{equation}
\vspace{-1.0em}
\begin{equation}
\centering
	\label{F1} 
	\begin{split}
 F1 = \frac{2 \times Precision \times Recall}{Precision + Recall}
	\end{split}
\end{equation}
\vspace{-1.0em}

\subsection{Experimental Settings}
To comprehensively evaluate our proposed model, all related experiments are implemented in PyTorch 1.8.1 (CUDA 11.1) on an NVIDIA GeForce RTX 3090 GPU with 24GB of memory. In the training phase, we selected the AdamW \cite{AdamW} optimizer and employed the cosine strategy to adjust the learning rate. Additionally, we utilized random horizontal and vertical flipping to augment the training data. According to the experimental settings in BuildFormer \cite{BuildFormer} and our hardware conditions, for the WHU building dataset, we set the initial learning rate to $10^{-3}$ and the batch size to 12. For the Massachusetts building dataset, we set the initial learning rate to $5e^{-4}$ and the batch size to 2. And for the Inria building dataset, we set the initial learning rate to $5e^{-4}$ and the batch size to 12.
 
\subsection{Compared Methods}
For a fair comparison, we selected two typical CNNs, $i.e.$, UNet\cite{UNet} based on VGG-16 \cite{VGG}, and HRNet\cite{HRNet} for the comparison. Meanwhile, we selected nine state-of-the-art deep learning methods designed for building extraction, $i.e.$, MA-FCN \cite{MA_FCN}, DSNet \cite{DSNet}, CBRNet\cite{CBRNet}, MSNet\cite{MSNet}, BOMSNet\cite{BOMSC-Net}, LCS\cite{LCS}, BuildFormer\cite{BuildFormer}, BCTNet \cite{BCTNet}, and FD-Net \cite{FD-Net}.

\subsection{Evaluation on WHU building dataset}
\subsubsection{Quantitative Comparison}
Table \ref{tab:table_test} lists the overall quantitative evaluation results of the different methods obtained on the WHU building dataset. Compared with other SOTA methods, our UANet can achieve the best performance on all metrics. In detail, our UANet outperforms SOTA method CBRNet ( \cite{CBRNet}) by $0.75$ percentage on the $IoU$ metric, $0.40$ percentage on the $F1$ metric, $0.65$ percentage on the $Precision$ metric, and $0.16$ percentage on the $Recall$ metric. The significant advantages of these metrics reflect the superiority of our method, proving that our proposed architecture with uncertainty consideration can greatly improve the effect of building extraction.

\begin{figure*}
\centering
\subfloat[Image ]{
\includegraphics[width=2.3cm,height=7.5cm]{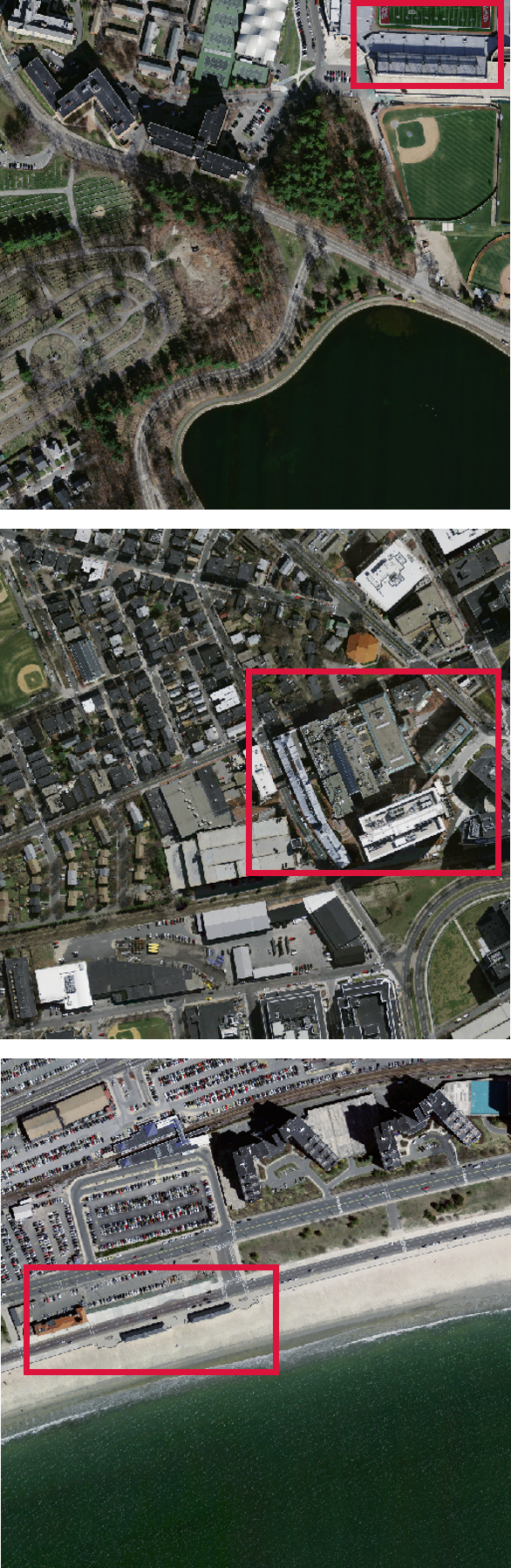}%
\label{fig_mass_tp_image}}
\hfil
\subfloat[GT]{
\includegraphics[width=2.3cm,height=7.5cm]{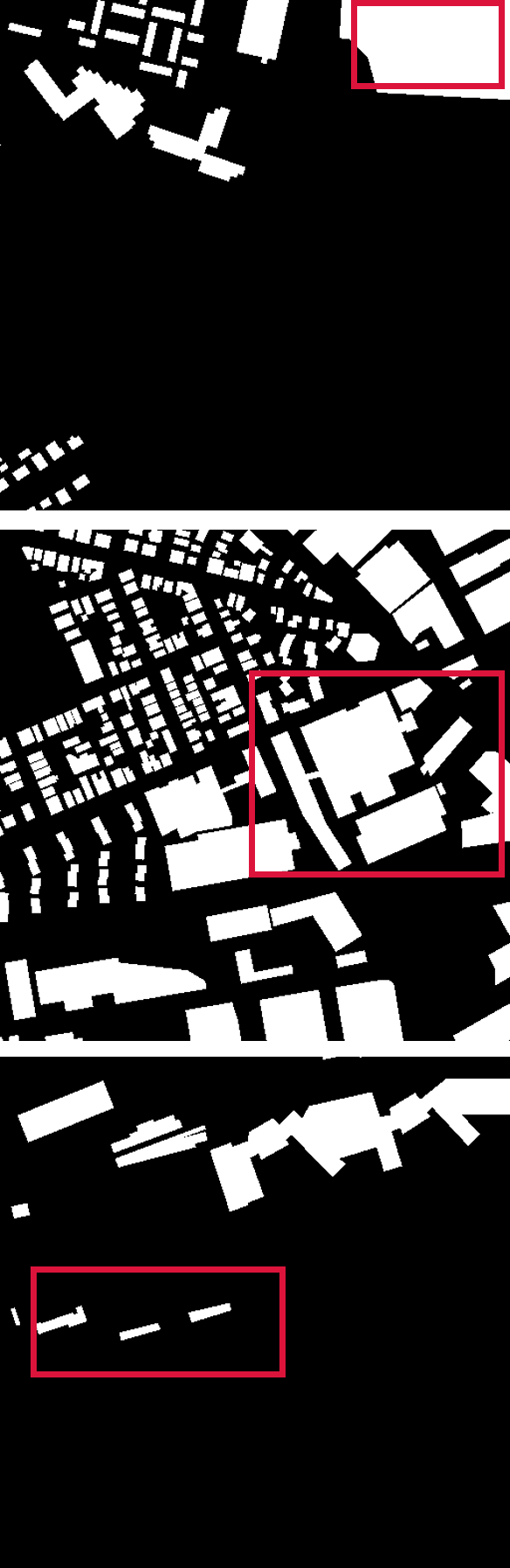}%
\label{fig_mass_tp_gt}}
\hfil
\subfloat[UNet]{
\includegraphics[width=2.3cm,height=7.5cm]{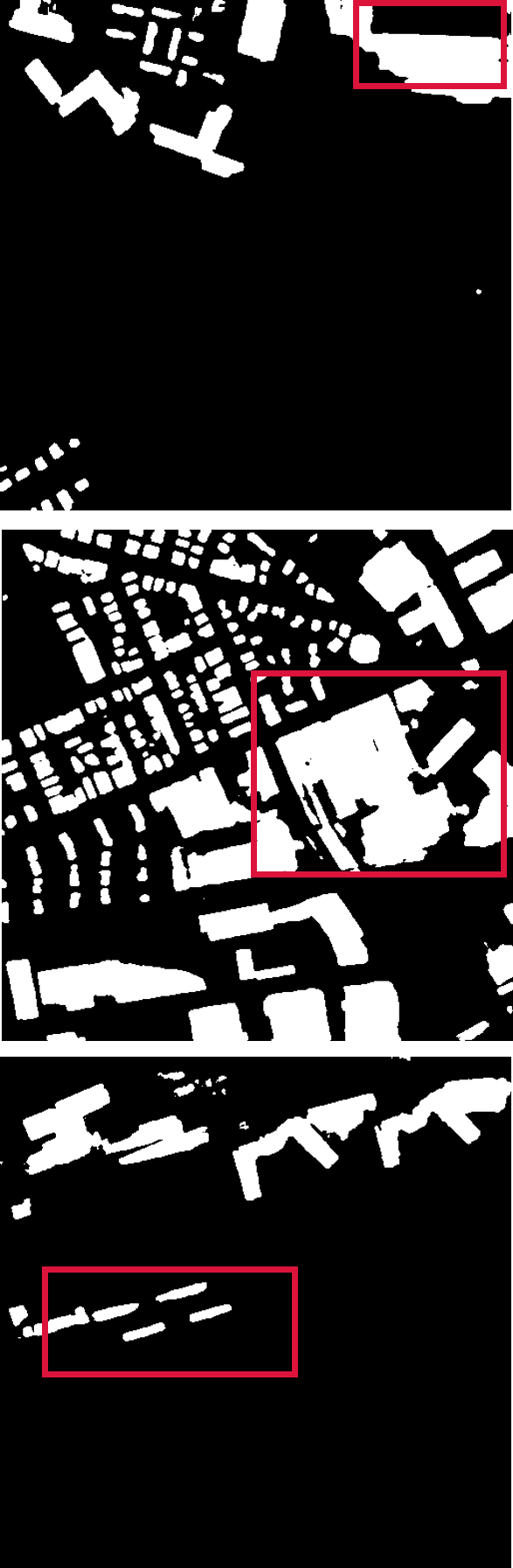}%
\label{fig_mass_tp_unet}}
\hfil
\subfloat[HRNet]{
\includegraphics[width=2.3cm,height=7.5cm]{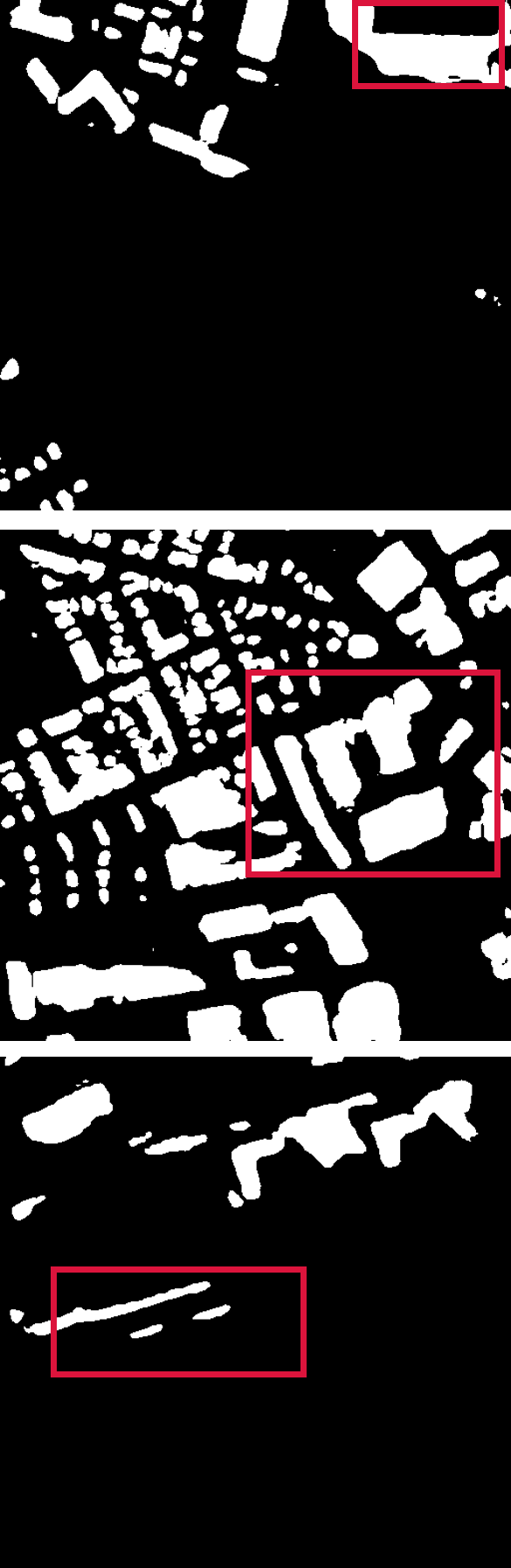}%
\label{fig_mass_tp_hrnet}}
\hfil
\subfloat[DSNet]{
\includegraphics[width=2.3cm,height=7.5cm]{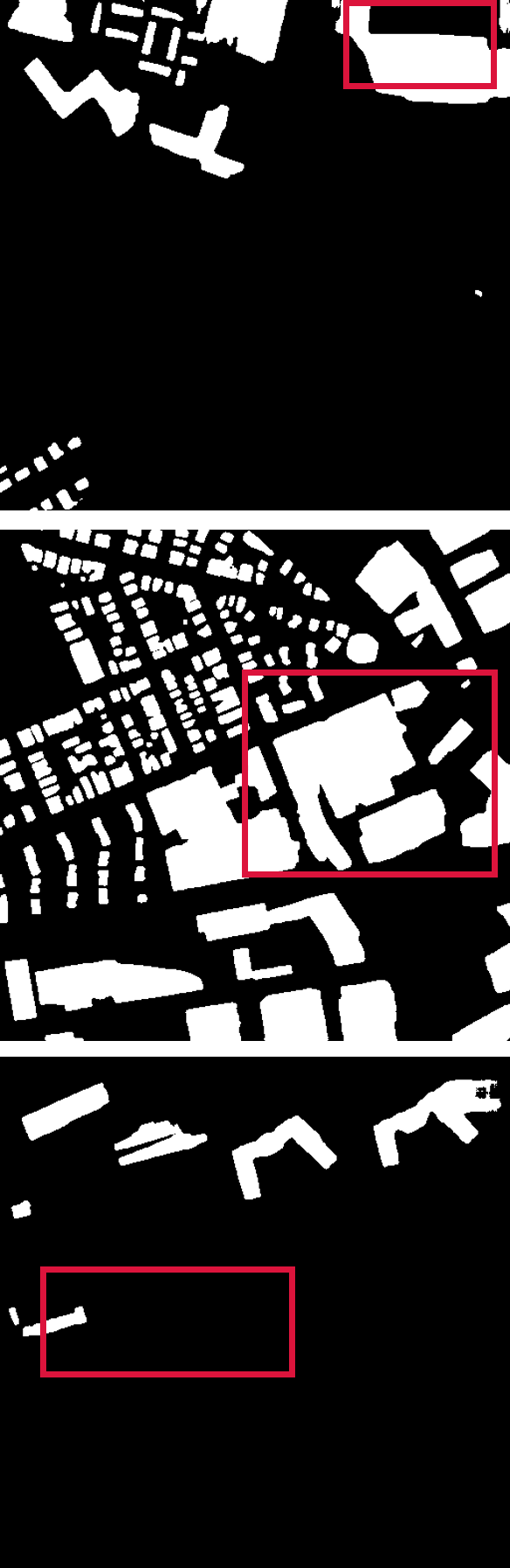}%
\label{fig_mass_tp_dsnet}}
\hfil
\subfloat[BuildFormer]{
\includegraphics[width=2.3cm,height=7.5cm]{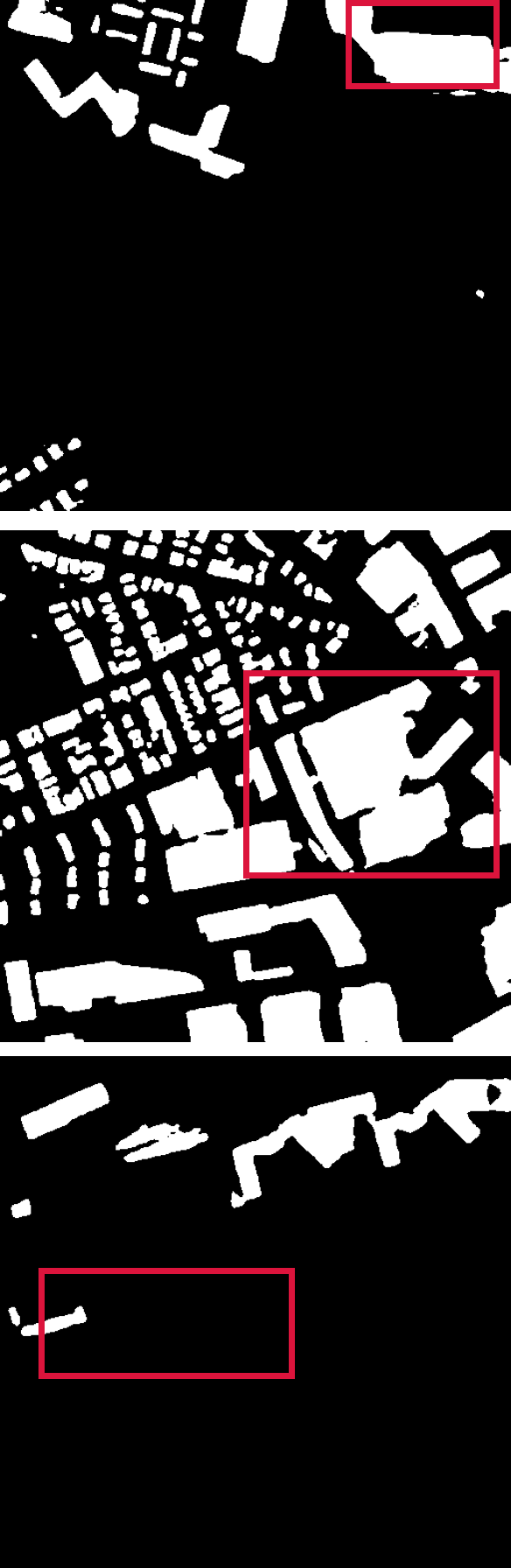}%
\label{fig_mass_tp_buildformer}}
\hfil
\subfloat[\textbf{Ours}]{
\includegraphics[width=2.3cm,height=7.5cm]
{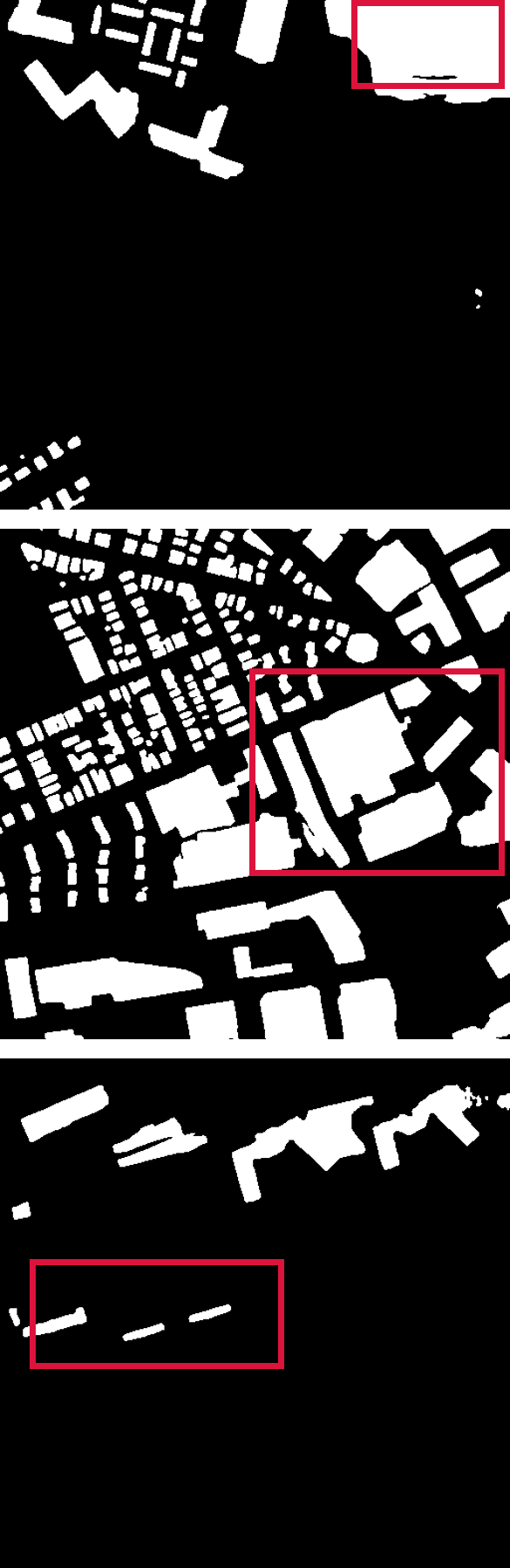}%
\label{fig_mass_tp_ours}}
\hfil
\caption{Visual Comparison on Massachusetts building dataset.}
\label{fig_tp_mass}
\vspace{-0.5em}
\end{figure*}

\subsubsection{Visual Comparison}

In order to compare our UANet with other SOTA methods more intuitively, we visualize the extraction results of all methods. As shown in Fig. \ref{fig_tp_whu}, the qualitative results for UANet and the other methods on the WHU buildings dataset are presented. For the first image, UNet, Deeplabv3+, HRNet, and BuildFormer all fail to extract the building in the red circle, while DSNet performs slightly better. By contrast, our UANet can accurately extract the buildings in the pink circle, which is closer to the ground truth. For the second image, all the compared methods wrongly recognize the road in the red circle as the part of buildings, but our UANet avoids this problem perfectly. Finally, for the third image, all the compared methods ignore the small building in the red circle, but our UANet demonstrates its superiority over the compared methods and successfully extracts this small building. It is evident that the ignored buildings in the three examples above are in a complex background, which leads to the uncertainty of the model. Faced with such a situation, our UANet is able to achieve satisfactory results with less uncertainty.

\subsection{Evaluation on Massachusetts building dataset}
\subsubsection{Quantitative Comparison}
Table \ref{tab:table_test} lists the overall quantitative evaluation results of the different methods obtained on the Massachusetts building dataset. Compared with other SOTA methods, our UANet can achieve the best performance on all metrics. Specifically, compared with the SOTA method DSNet, our UANet can outperform it by $1.37$ percentage on the IoU metric, $0.89$ percentage on the F1 metric, $0.37$ percentage on the Precision metric, and $0.56$ percentage on the Recall metric. Since the same backbone is used as other compared methods (except BuildFormer), the huge advantage of our UANet indicates that our decoding strategy is very effective.

\subsubsection{Visual Comparison}
As shown in Fig. \ref{fig_tp_mass}, we present three visual examples of all the compared methods and our UANet on the Massachusetts building dataset. Due to the low image resolution of the dataset and the dense distribution of buildings in the image, it is evident that all the methods have a lot of errors in their extraction results. However, it is obvious that our extraction result extracts more details such as texture and edge than the compared methods, which is most noticeable in the red box area. The more complex the environment, the better our UANet performs than other compared methods, as our UANet can highlight the uncertain areas and eliminate them to a large extent.

\subsection{Evaluation on Inria building dataset}
\subsubsection{Quantitative Comparison}
As shown in Table \ref{tab:table_test}, we list the overall quantitative evaluation results of the different methods tested on the Inria building dataset. Compared with the SOTA method BuildFormer, it is clear that our UANet can outperform it by $1.84$ percentage on $IoU$, $1.05$ percentage on $F1$, $1.39$ percentage on $Precison$, and $0.74$ percentage on $Recall$. This significant improvement demonstrates the effectiveness of our approach to introduce uncertainty to optimize decoding strategies.

\begin{figure*}
\centering
\subfloat[Image ]{
\includegraphics[width=2.3cm,height=7.5cm]{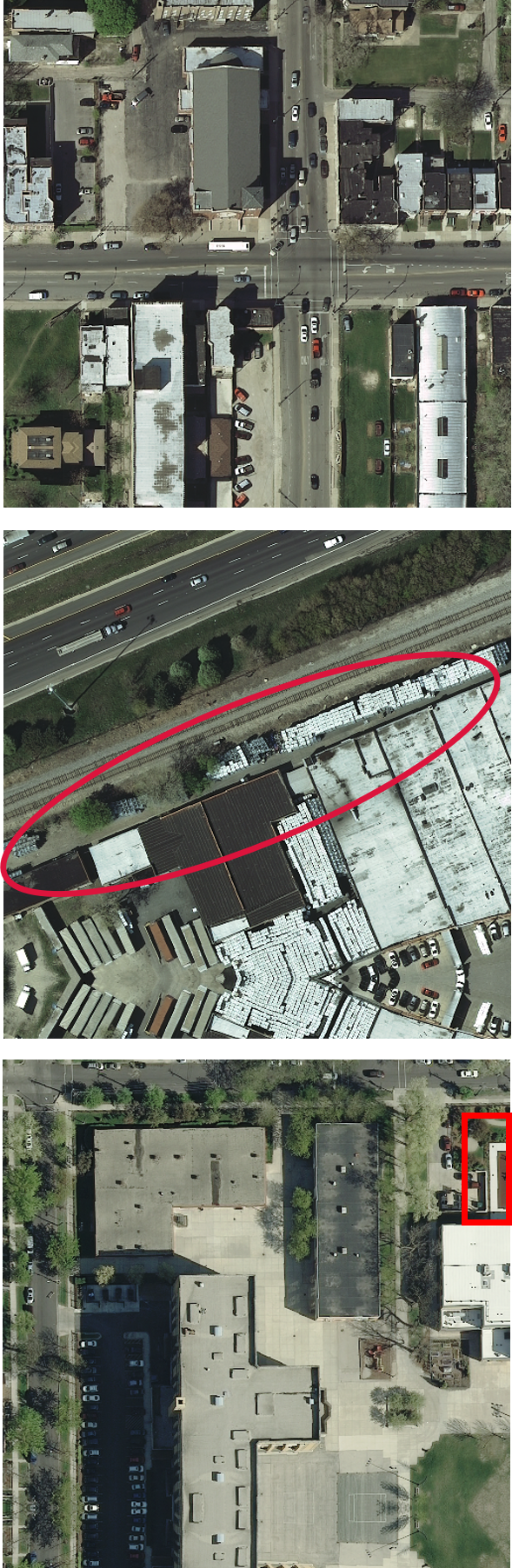}%
\label{fig_inria_tp_image}}
\hfil
\subfloat[GT]{
\includegraphics[width=2.3cm,height=7.5cm]{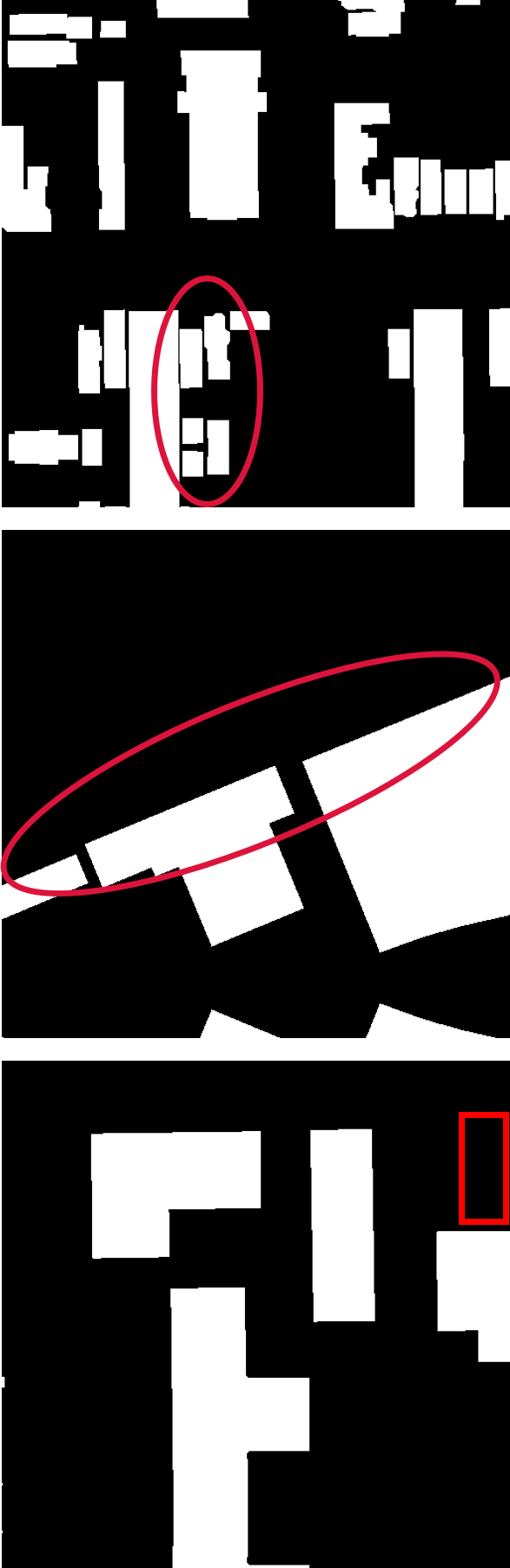}%
\label{fig_inria_tp_gt}}
\hfil
\subfloat[UNet]{
\includegraphics[width=2.3cm,height=7.5cm]{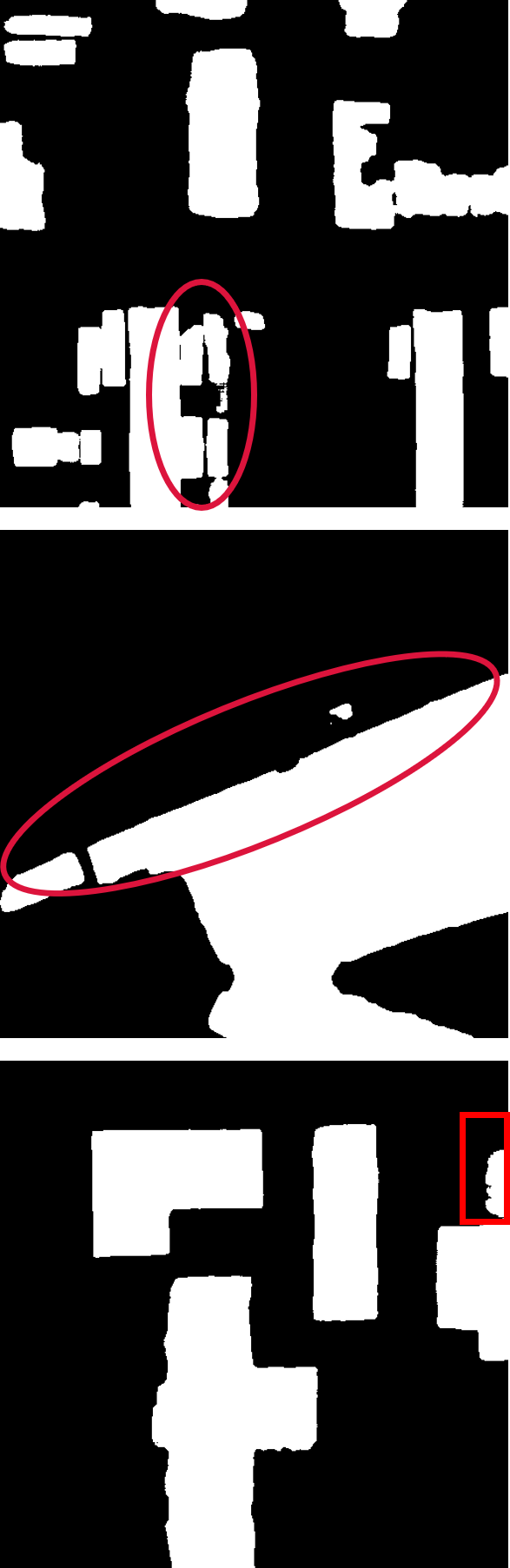}%
\label{fig_inria_tp_unet}}
\hfil
\subfloat[HRNet]{
\includegraphics[width=2.3cm,height=7.5cm]{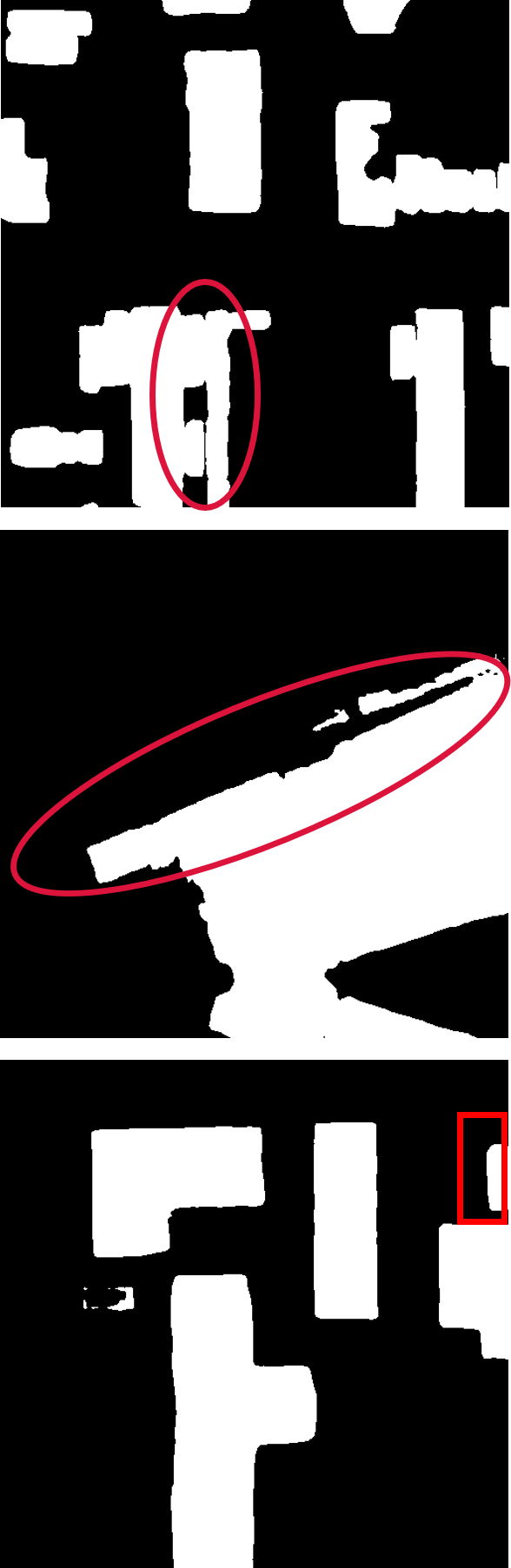}%
\label{fig_inria_tp_hrnet}}
\hfil
\subfloat[DSNet]{
\includegraphics[width=2.3cm,height=7.5cm]{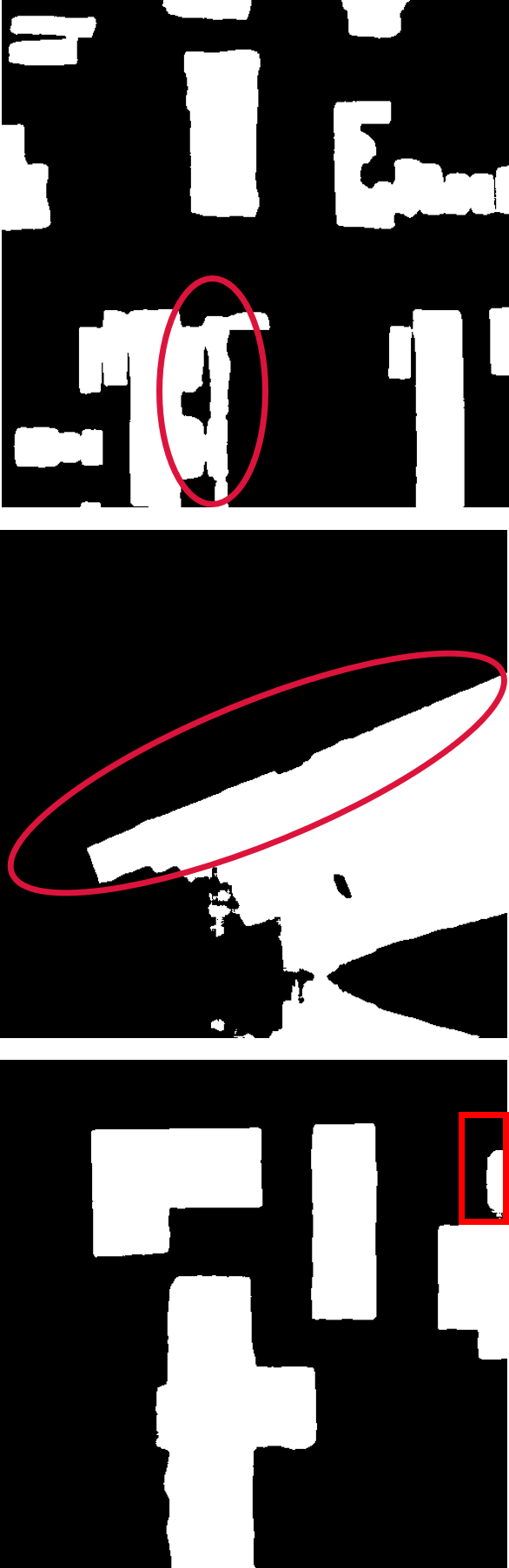}%
\label{fig_inria_tp_dsnet}}
\hfil
\subfloat[BuildFormer]{
\includegraphics[width=2.3cm,height=7.5cm]{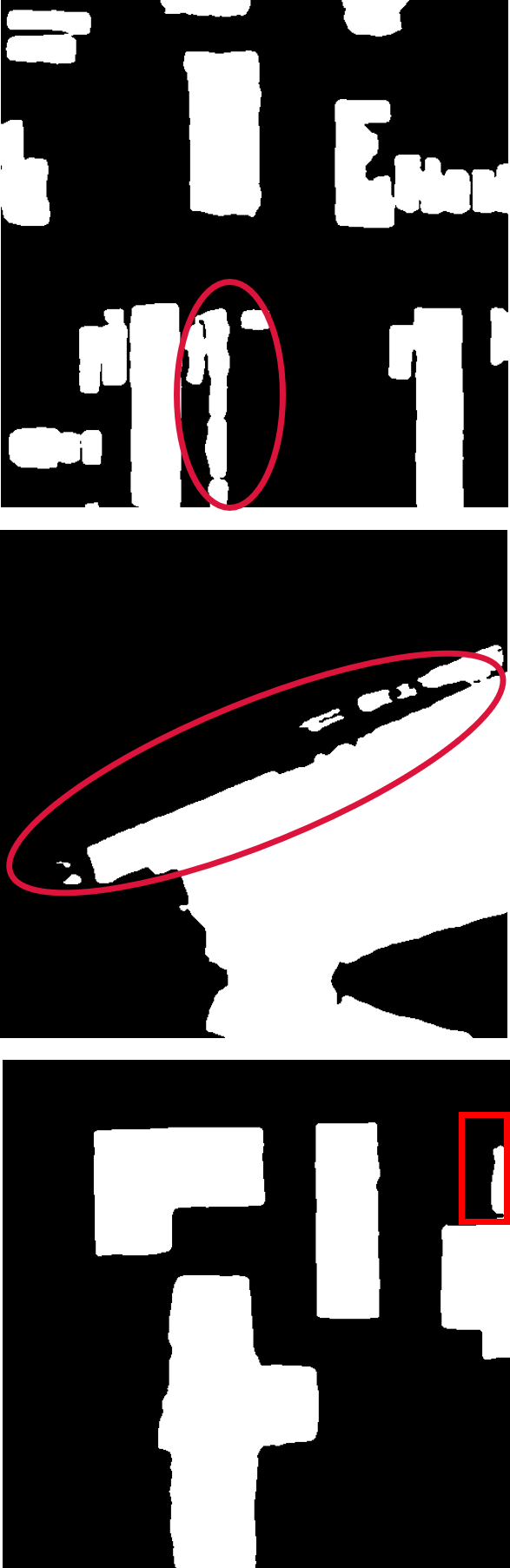}%
\label{fig_inria_tp_buildformer}}
\hfil
\subfloat[\textbf{Ours}]{
\includegraphics[width=2.3cm,height=7.5cm]
{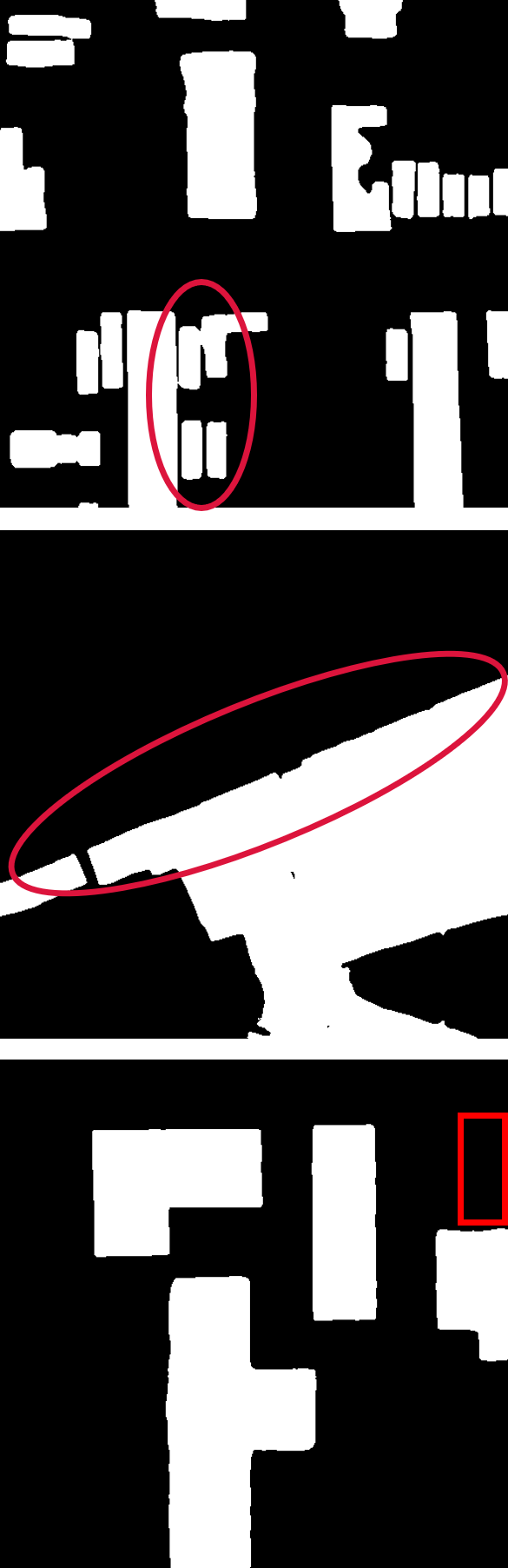}%
\label{fig_inria_tp_ours}}
\hfil
\caption{Visual Comparison on Inria building dataset.}
\label{fig_tp_inria}
\vspace{-1.0em}
\end{figure*}

\subsubsection{Visual Comparison}
As presented in Fig. \ref{fig_tp_inria}, we select three typical examples to compare our UANet with the other SOTA methods. In the first image, we can see that the buildings in the red circle is covered by shadows from the buildings next to it, and the compared methods fail to successfully extract the whole bodies of the buildings.
At the same time, compared with our result, there are still more drawbacks. In the second image, we can easily find that the buildings in the red circle are somewhat different from the other buildings around it, and HRNet, DSNet, and BuildFormer ignore the real part of buildings but mistake unrelated parts for buildings. By contrast, the result of our proposed UANet is very close to the ground truth. In the third image, it is easy to find that the compared methods mistakenly detect the part of the red rectangle as a building, but our UANet succeed. From these three examples, it is convinced that our UANet can make the right judgment in the face of complex environments.

\section{Ablation Study}
In order to explore the effectiveness of our proposed modules in UANet, we conduct extensive experiments on the three building datasets. We selected the general encoder-decoder network used in our UANet as the baseline, which utilizes VGG-16 as the encoder and use a conventional decoding method to output an uncertain extraction map. Based on it, we verify the effectiveness of the Prior Information Guide Module (PIGM), and the Uncertainty-Aware Fusion Module (UAFM) in turn. In the following parts, we will give a detailed analysis. 

\begin{table*}[]
\normalsize
\setlength\tabcolsep{3pt}
\caption{Ablation results on the test dataset.}
\label{tab:table_abn}
\center
\begin{tabular}{ccc|cccc|cccc|cccc}
\hline
\multirow{2}{*}{Baseline} &\multirow{2}{*}{PIGM} &\multirow{2}{*}{UAFM} & \multicolumn{4}{c|}{WHU (\%)}&\multicolumn{4}{c|}{Massachusetts (\%)} &\multicolumn{4}{c}{Inira (\%)}                                             
 \\ \cline{4-15} 
 & & &$IoU\uparrow$ & $F1\uparrow$ & $Pre\uparrow$ & $Recall\uparrow$ 
 & $IoU\uparrow$ & $F1\uparrow$ & $Pre\uparrow$ & $Recall\uparrow$
 & $IoU\uparrow$ & $F1\uparrow$ & $Pre\uparrow$ & $Recall\uparrow$ \\ \hline
\checkmark & & &
87.35  &92.90 &92.25 &93.56
&69.73 &82.17 &85.41 &79.16
&79.08 &88.32 &87.77 &88.88\\ 
\checkmark & & \checkmark &
\underline{91.25} &\underline{95.43} &\underline{95.73} &\underline{95.13}
&\underline{74.84} &\underline{85.61} &\underline{87.56} &\underline{83.75}
&\underline{81.84} &\underline{90.01} &\underline{90.43} &\underline{89.61}\\ 
\checkmark & \checkmark & \checkmark&
\color{red}\textbf{92.15} &{\color{red}\textbf{95.91}} &{\color{red}\textbf{95.96}} &\color{red}\textbf{95.86}

& \color{red}\textbf{{76.41}} &\color{red}\textbf{{86.63}} &\color{red}\textbf{87.94} &\color{red}\textbf{{85.35}}

& {\color{red}\textbf{83.08}} &{\color{red}\textbf{90.76}} &{\color{red}\textbf{92.04}} &{\color{red}\textbf{89.52}}
\\ \hline
\end{tabular}
\end{table*}

\subsection{The effectiveness of PIGM}
\begin{table*}[]
\normalsize
\setlength\tabcolsep{3pt}
\caption{The ablation results about PIGM on the test dataset.}
\label{tab:table_PIGM}
\center
\begin{tabular}{cccc|cccc|cccc|cccc}
\hline
\multirow{2}{*}{Baseline} &\multirow{2}{*}{SC} &\multirow{2}{*}{CC} & \multirow{2}{*}{UAFM} &\multicolumn{4}{c|}{WHU (\%)}&\multicolumn{4}{c|}{Massachusetts (\%)} &\multicolumn{4}{c}{Inira (\%)}                                             
 \\ \cline{5-16} 
 & & & & $IoU\uparrow$ & $F1\uparrow$ & $Pre\uparrow$ & $Recall\uparrow$ 
 & $IoU\uparrow$ & $F1\uparrow$ & $Pre\uparrow$ & $Recall\uparrow$
 & $IoU\uparrow$ & $F1\uparrow$ & $Pre\uparrow$ & $Recall\uparrow$ \\ \hline
\checkmark & & & \checkmark&
\underline{91.25} &\underline{95.43} &\underline{95.73} &95.13
&74.84 &85.61 &87.56 &83.75
&81.84 &90.01 &90.43 &\color{red}\textbf{89.61}\\ 
\checkmark & \checkmark & & \checkmark&
91.02 &95.27 &95.22 &95.32
&75.03 &85.92 &{87.78} &84.14
&82.61 &90.48 &\underline{91.68} &89.30\\
\checkmark & &\checkmark & \checkmark&
91.15 &95.28 &95.23 &\underline{95.34}
&\underline{75.07} &\underline{85.98} &\underline{87.84} &\underline{84.20}
&\underline{82.78} &\underline{90.58} &91.66 &\underline{89.52}\\ 
\checkmark &\checkmark &\checkmark & \checkmark&
\color{red}\textbf{92.15} &{\color{red}\textbf{95.91}} &{\color{red}\textbf{95.96}} &\color{red}\textbf{95.86}

& \color{red}\textbf{{76.41}} &\color{red}\textbf{{86.63}} &\color{red}\textbf{87.94} &\color{red}\textbf{{85.35}}

& {\color{red}\textbf{83.08}} &{\color{red}\textbf{90.76}} &{\color{red}\textbf{92.04}} &{\underline{89.52}}
\\ \hline
\end{tabular}
\end{table*}

Guided by the uncertain extraction map $M_5$, we try to enhance the highest-level features via PIGM. Different from the previous attention mechanism, we introduced a cross-attention method, which helps the high dimensions features to learn the spatial and the semantic relationship channel by channel. As shown in Table \ref{tab:table_abn}, by introducing the PIGM, the extraction accuracy can be significantly improved. We conducted several experiments to verify the detailed effect of the two components of the PIGM. As shown in Table \ref{tab:table_PIGM}, to verify the effectiveness of PIGM, we conducted four sets of experiments: 1) without learning any correlation 2) just establishing the spatial correlation (SC), 3) just establishing the channel correlation (CC), 4) establishing the spatial and channel correlation in series (SC + CC). It is clear that the two enhancement ways played their own role in the PIGM module.

\subsection{The effectiveness of UAFM}

The proposed UAFM can reduce the uncertainty of $G_{i}$ with the help of
the foreground uncertainty rank map $R_{f}^{i}$ and the background uncertainty rank map $R_{b}^{i}$, and output feature $G_{i-1}$ with lower uncertainty. As shown in Table \ref{tab:table_abn}, we can easily find that the UAFM can bring significant improvement of building extraction performance.
\par


We also conducted extensive experiments to explore the accuracy improvement brought about by such an uncertainty-aware strategy in detail. As presented in Table \ref{tab:table_UAFM}, We set up four feature interaction methods: \textbf{$Case 1$}: just concatenate the adjacent layers of features and introduce deep supervision in all levels; \textbf{$Case 2$}: just use the $Sigmoid$ function to process the extraction map from former level and utilize it achieve the feature interaction; \textbf{$Case 3$}: just use the foreground uncertainty ($R_{f}^{i}$) to achieve the feature interaction; \textbf{$Case 4$}: follow our proposed uncertainty-aware strategy which utilizes both the foreground uncertainty ($R_{f}^{i}$) and the background uncertainty ($R_{b}^{i}$) to achieve the feature interaction. 
It is evident that the extraction accuracy is significantly improved with the guidance of the uncertainty maps of both the foreground and the background, which can intuitively reflect the huge advantage of our proposed strategy.

\begin{table*}[]
\normalsize
\setlength\tabcolsep{3pt}
\caption{The ablation results about UAFM on the test dataset.}
\label{tab:table_UAFM}
\center
\begin{tabular}{c|cccc|cccc|cccc}
\hline
\multirow{2}{*}{Baseline} & \multicolumn{4}{c|}{WHU (\%)}&\multicolumn{4}{c|}{Massachusetts (\%)} &\multicolumn{4}{c}{Inira (\%)}                                             
 \\ \cline{2-13} 
 & $IoU\uparrow$ & $F1\uparrow$ & $Pre\uparrow$ & $Recall\uparrow$ 
 & $IoU\uparrow$ & $F1\uparrow$ & $Pre\uparrow$ & $Recall\uparrow$
 & $IoU\uparrow$ & $F1\uparrow$ & $Pre\uparrow$ & $Recall\uparrow$ \\ \hline
$Case 1$&
89.08	&94.23	&93.92	&94.53	
&72.13	&83.81	&86.56	&81.24 
&79.73  &88.73  &89.17  &88.28\\
$Case 2$&
91.39 &95.43 &\underline{95.47} &95.40
&75.21 &85.96 &{87.81} &84.18 
&80.98 &89.03 &90.62 &87.50\\ 
$Case 3$ &
\underline{91.61} &\underline{95.57} &95.45 &\underline{95.70}
&\underline{75.87} &\underline{86.28} &\underline{87.93} &\underline{84.69}
&\underline{82.34} &\underline{90.31} &\underline{91.49} &\underline{89.16}\\  
$Case 4$ &
\color{red}\textbf{92.15} &{\color{red}\textbf{95.91}} &{\color{red}\textbf{95.96}} &\color{red}\textbf{95.86}

& \color{red}\textbf{{76.41}} &\color{red}\textbf{{86.63}} &\color{red}\textbf{87.94} &\color{red}\textbf{{85.35}}

& {\color{red}\textbf{83.08}} &{\color{red}\textbf{90.76}} &{\color{red}\textbf{92.04}} &{\color{red}\textbf{89.52}}
\\ \hline
\end{tabular}
\end{table*}

\begin{table*}[]
\center
\normalsize
\setlength\tabcolsep{3pt}
\caption{The results about $M_{i}$ on the test dataset.}
\label{tab:table_M}
\begin{tabular}{c|cccc|cccc|cccc}
\hline
\multirow{2}{*}{Baseline} & \multicolumn{4}{c|}{WHU (\%)}&\multicolumn{4}{c|}{Massachusetts (\%)} &\multicolumn{4}{c}{Inira (\%)}                                             
 \\ \cline{2-13} 
 & $IoU\uparrow$ & $F1\uparrow$ & $Pre\uparrow$ & $Recall\uparrow$ 
 & $IoU\uparrow$ & $F1\uparrow$ & $Pre\uparrow$ & $Recall\uparrow$
 & $IoU\uparrow$ & $F1\uparrow$ & $Pre\uparrow$ & $Recall\uparrow$ \\ \hline
$M_{5}$&
87.35  &92.90 &92.25 &93.56
&69.73 &82.17 &85.41 &79.16
&79.08 &88.32 &87.77 &88.88\\
$M_{4}$&
89.93 &94.70 &95.16 &94.24
&74.52 &85.40 &86.46 &84.37
&82.09 &90.16 &91.45 &88.91\\ 
$M_{3}$ &
91.25 &95.43 &95.94 &94.92
&75.90 &86.30 &87.17 &\underline{85.45}
&82.83 &90.61 &91.95 &89.31\\ 
$M_{2}$ &
\underline{91.68} &\underline{95.66} &\color{red}\textbf{96.16} &\underline{95.17}
&\underline{76.10} &\underline{86.43} &\underline{87.41} &\color{red}\textbf{85.47}
&\underline{83.05} &\underline{90.74} &\underline{92.02} &\underline{89.50}\\ 
$M_{1}$ &
\color{red}\textbf{92.15} &{\color{red}\textbf{95.91}} &{\underline{95.96}} &\color{red}\textbf{95.86}

& \color{red}\textbf{{76.41}} &\color{red}\textbf{{86.63}} &\color{red}\textbf{87.94} &{85.35}

& {\color{red}\textbf{83.08}} &{\color{red}\textbf{90.76}} &{\color{red}\textbf{92.04}} &{\color{red}\textbf{89.52}}
\\ \hline
\end{tabular}
\vspace{1.0em}
\end{table*}

\begin{figure*}
\centering
\subfloat[Image/GT ]{
\includegraphics[width=2.8cm,height=5.5cm]{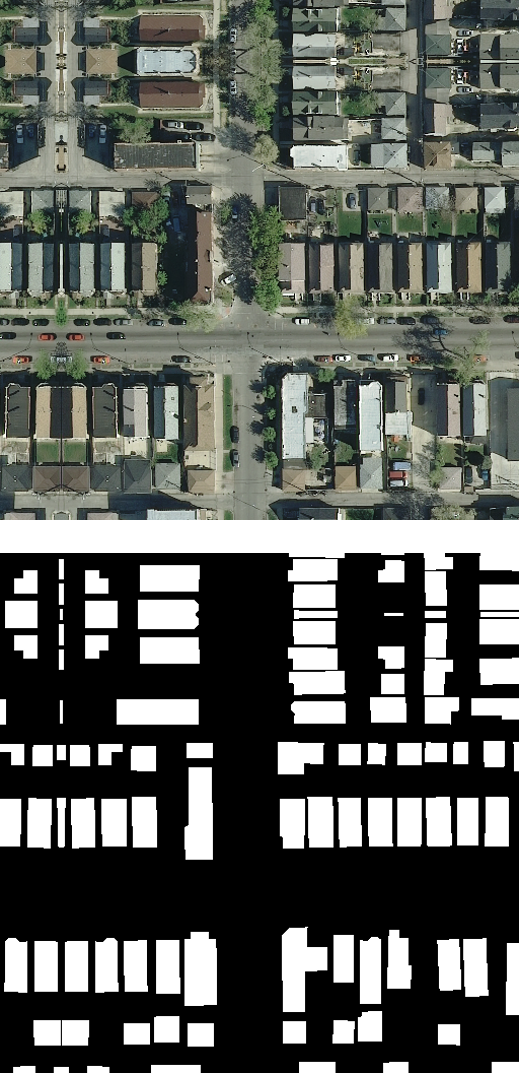}%
\label{fig_uafm_img_gt}}
\hfil
\subfloat[5st level]{
\includegraphics[width=2.8cm,height=5.5cm]{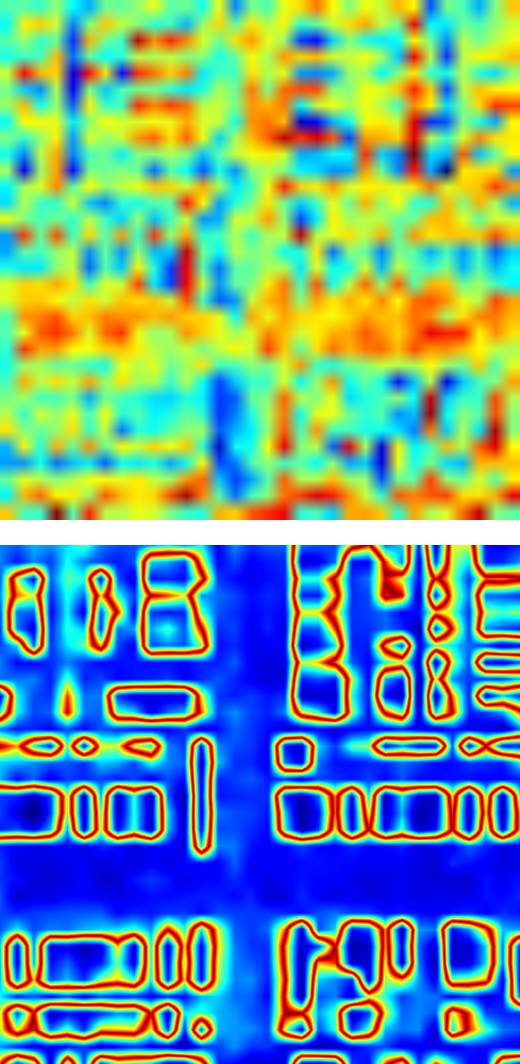}%
\label{fig_uafm_1}}
\hfil
\subfloat[4st level]{
\includegraphics[width=2.8cm,height=5.5cm]{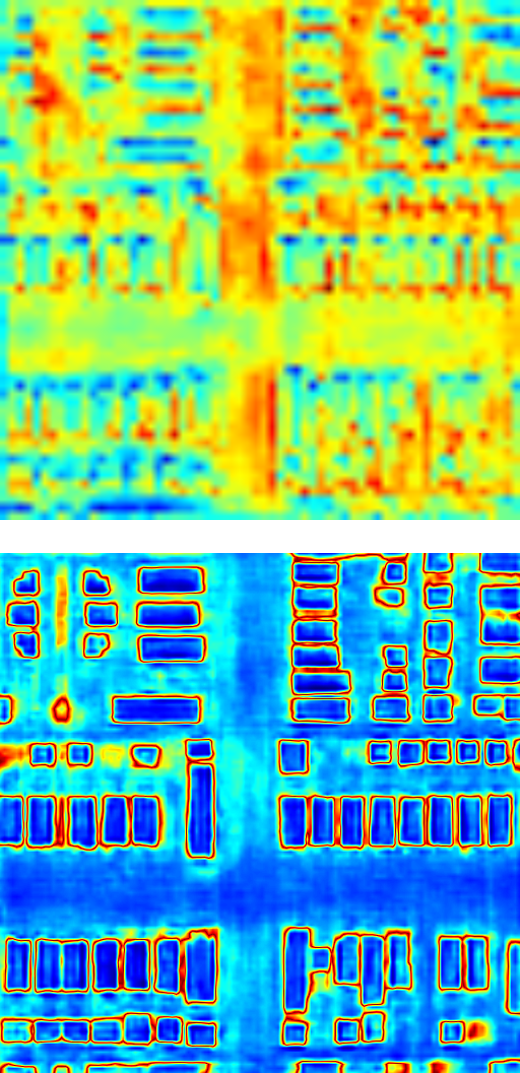}%
\label{fig_uafm_2}}
\hfil
\subfloat[3st level]{
\includegraphics[width=2.8cm,height=5.5cm]{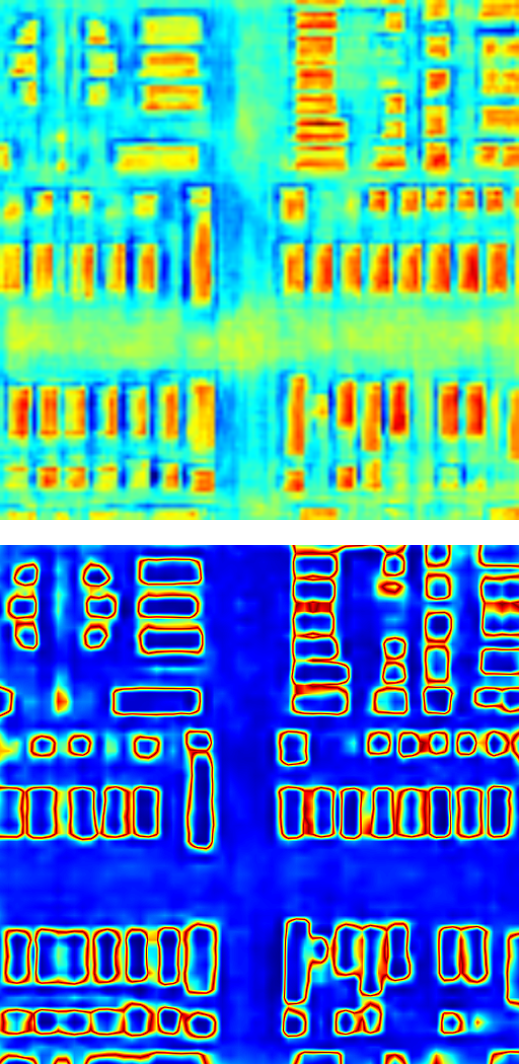}%
\label{fig_uafm_3}}
\hfil
\subfloat[2st level]{
\includegraphics[width=2.8cm,height=5.5cm]{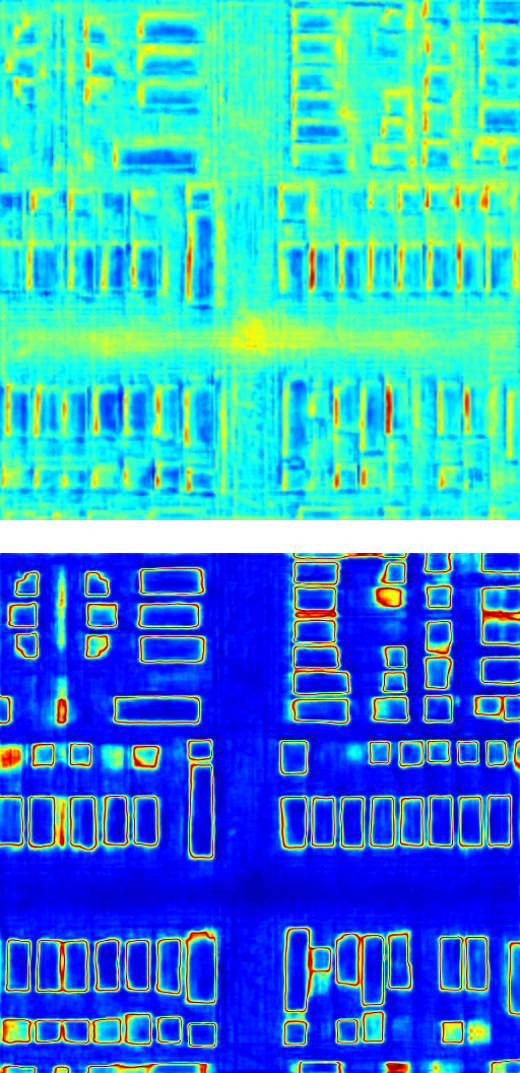}%
\label{fig_uafm_4}}
\hfil
\subfloat[1st level]{
\includegraphics[width=3.0cm,height=5.5cm]{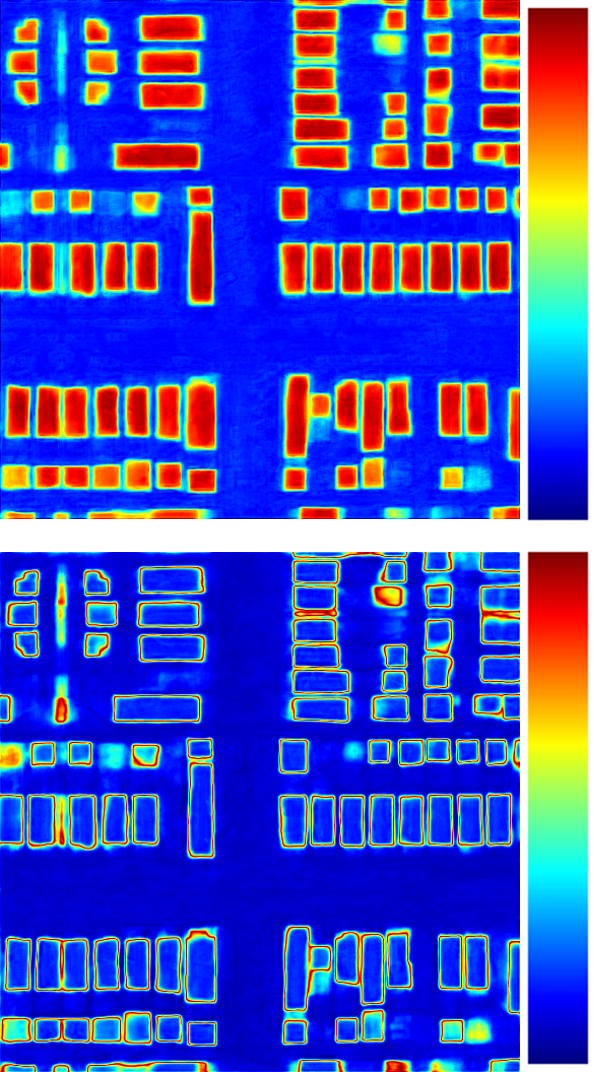}%
\label{fig_uafm_5}}
\hfil
\caption{Visual examples of building extraction. The first row represents the visualizations of $G_{i}$, and the second row represents the uncertainty visualization of $M_{i}$.}
\label{fig_uafm_abalation}
\vspace{-0.5em}
\end{figure*}

At the same time, In order to verify that UAFM can output feature $G_{i-1}$ and related prediction $M_{i-1}$ with lower uncertainty, we visualize $G_{i-1}$ and the uncertainty reflected in $M_{i-1}$ of all levels.
As exhibited in Fig .\ref{fig_uafm_abalation}, we can observe that in each level, the enhanced features $G_{i-1}$ can achieve cleaner objects and related edges compared to that of $G_{i}$, and the uncertainty is progressively reduced. Besides, Table. \ref{tab:table_M} can also illustrate the gradual enhancement of our high-to-low uncertain-aware strategy from quantitative evaluation.

\subsection{The analysis of URA}

\begin{figure*}
\centering
\subfloat[Image/GT ]{
\includegraphics[width=2.6cm,height=5.5cm]{UAFM/img.png}%
\label{fig_ura_img_gt}}
\hfil
\subfloat[5st level]{
\includegraphics[width=2.6cm,height=5.5cm]{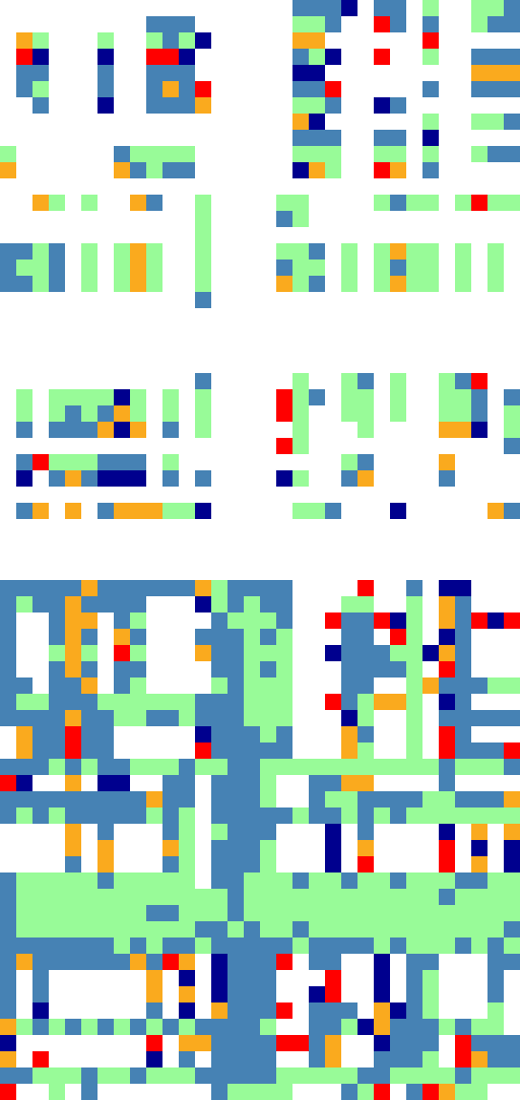}%
\label{fig_ura_5}}
\hfil
\subfloat[4st level]{
\includegraphics[width=2.6cm,height=5.5cm]{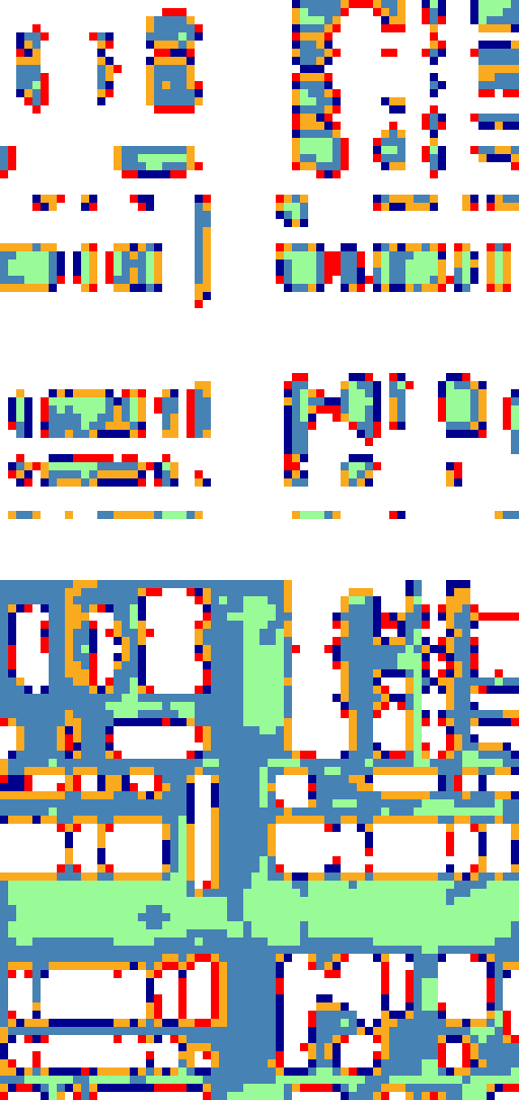}%
\label{fig_ura_4}}
\hfil
\subfloat[3st level]{
\includegraphics[width=2.6cm,height=5.5cm]{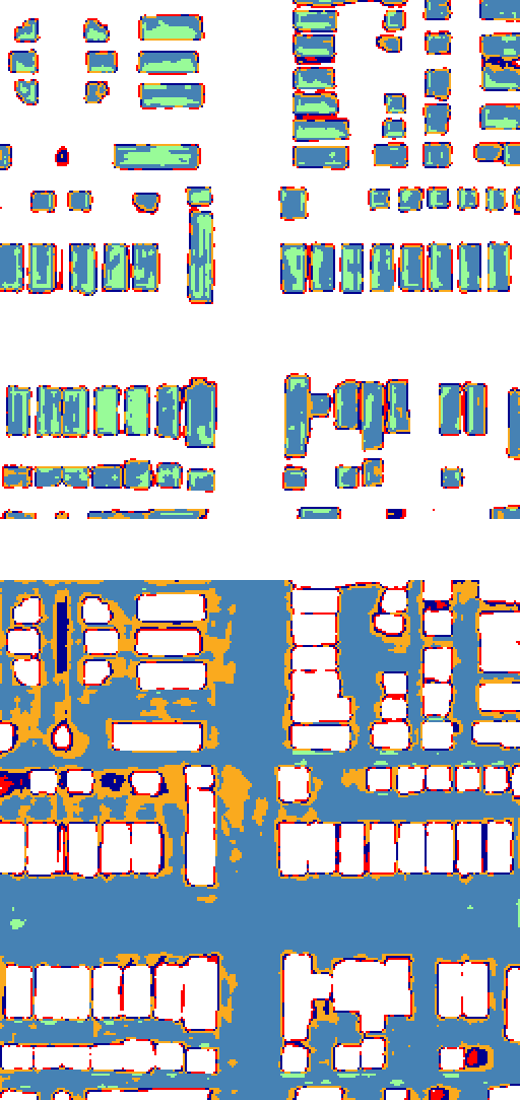}%
\label {fig_ura_3}}
\hfil
\subfloat[2st level]{
\includegraphics[width=2.6cm,height=5.5cm]{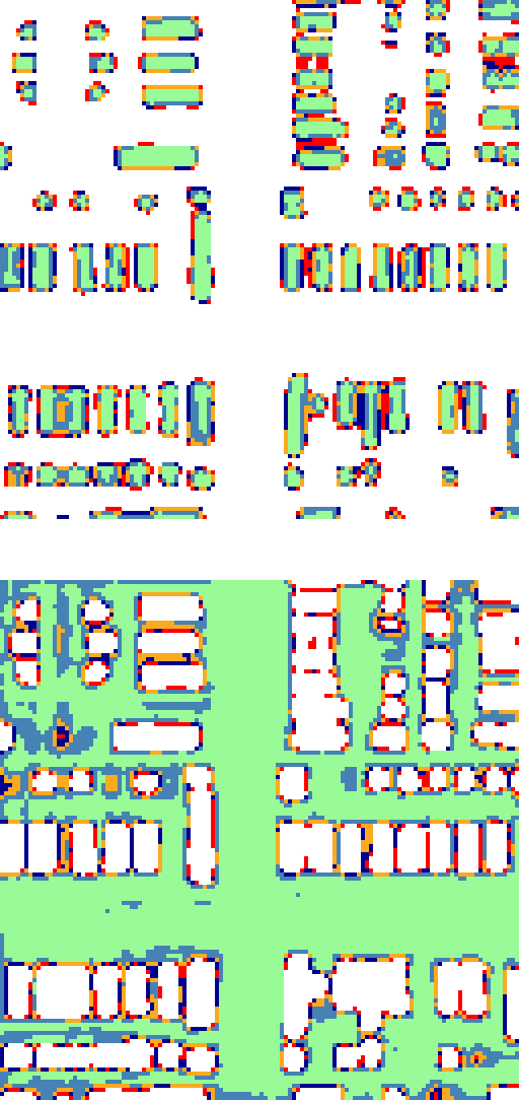}%
\label{fig_ura_2}}
\hfil
\subfloat[1st level]{
\includegraphics[width=4.0cm,height=5.5cm]{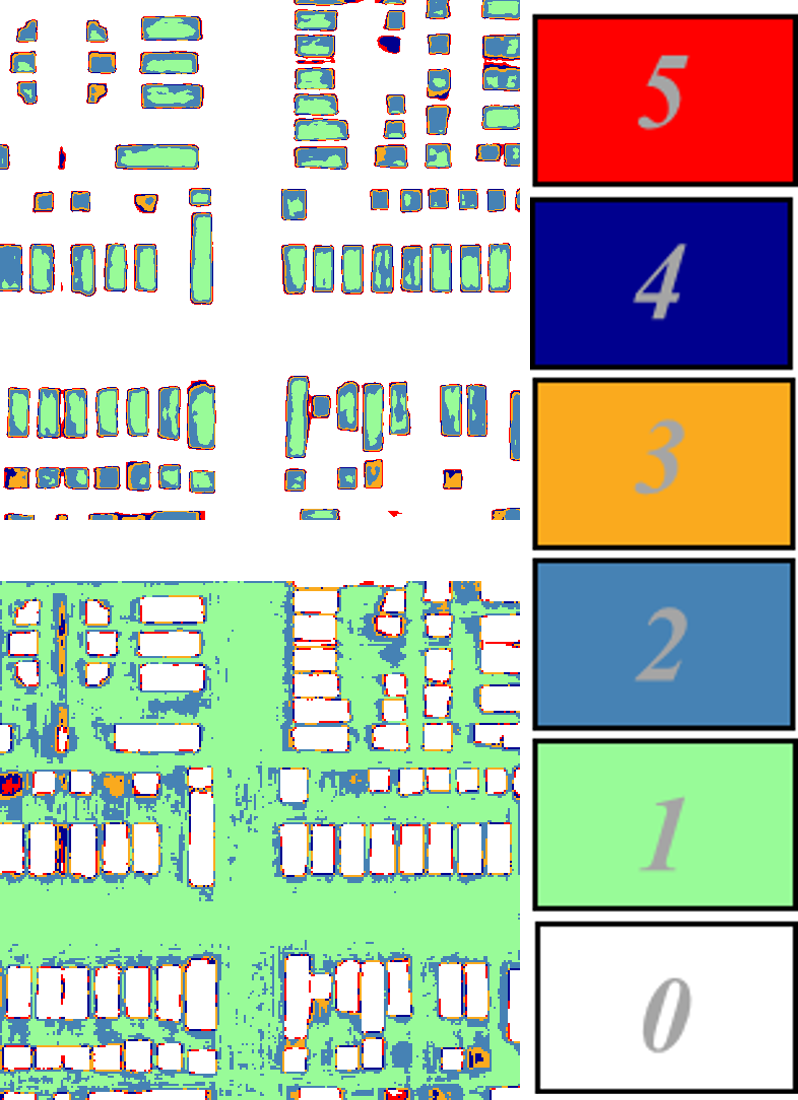}%
\label{fig_ura_1}}
\hfil
\caption{Visual examples of building extraction. The first row represents the visualizations of $R_{f}^{i}$, and the second row represents the uncertainty visualization of $R_{b}^{i}$.}
\label{fig_URA_abalation}
\vspace{-0.5em}
\end{figure*}

As the key algorithm in our UAFM, URA aims to rank the uncertainty level of all pixels in the extraction map. As mentioned in Section \ref{sec:UAFM}, the principle of URA is to define a non-increasing linear function $\mathbb{f}$ from $U$ to $R$. To simplify our design of $\mathbb{f}$, we define the uncertainty of $0-0.5$ into five levels. To verify the effectiveness of our designed URA, we visualize both $R_{f}^{i}$ and $R_{b}^{i}$ ($\left\{i=1,2,3,4,5 \right \}$). As shown in Fig. \ref{fig_URA_abalation}, we find that the level of uncertainty is decreasing overall. We can conclude that assigning different weights to each level of uncertainty can address the uncertainty problem to some extent. We also find that, the weight of pixels with high uncertainty needs to be significantly higher than that of pixels with low uncertainty. 

\subsection{The analysis of different encoders}
\begin{table}[]
\normalsize
\setlength\tabcolsep{3pt}
\caption{Ablation results about different encoders on the test dataset.}
\label{tab:table_Encoder}
\center
\begin{tabular}{c|cccc}
\hline
\multirow{2}{*}{Method} &\multicolumn{4}{c}{Inira (\%)}                                             
 \\ \cline{2-5} 
 & $IoU\uparrow$ & $F1\uparrow$ & $Pre\uparrow$ & $Recall\uparrow$ \\ \hline
ResNet-50&
82.17 &90.21 &91.24 &89.20\\
Res2Net-50&
\underline{83.17} &\underline{90.81} &\underline{91.89} &\underline{89.76}\\ 
PVT-V2-B2 &
\color{red}\textbf{83.34} &\color{red}\textbf{90.91} &91.86 &\color{red}\textbf{89.97}\\ 
VGG-16 &
{83.08}  &{90.76} &\color{red}\textbf{92.04} &{89.52}
\\ \hline
\end{tabular}
\end{table}
As mentioned before, our UANet can be also used for other kinds of encoder-decoder building extraction models to improve the certainty prediction. And we select ResNet-50 \cite{ResNet}, Res2Net-50 \cite{Res2Net}, VGG-16 \cite{VGG} and PVT-V2-B2 \cite{PVT} as encoder-decoder backbones, to testify the efficacy of our UANet. As illustrated in Table\ref{tab:table_Encoder}, we can easily find that our UANet can achieve excellent results on different encoders, especially in the case of transformer based architecture PVT-V2-B2. However, since most previous models utilize the VGG-16 as the backbone, we also choose the same setting for a fair comparison.


\subsection{The comparison with other uncertainty strategies}
In our proposed UANet, we rank the uncertainty-level from both the foreground and the background perspectives to reduce the uncertainty of features level by level. To verify the superiority over other uncertainty strategies, we compared our method with the uncertainty strategies used in other vision tasks. In detail, on the one hand, we adopted the settings in \cite{aleatoric_uncertainty} and added a confidence estimation network to our VGG-16 based general encoder-decoder structure to formalise the uncertainty as probability distribution over model output and the input image. On the other hand, we followed the setting in \cite{CVAE_COD} and introduced the Conditional Variational Autoencoder (CAVE) to measure the uncertainty of input data, which was followed by being input to our VGG-16 based general encoder-decoder structure with the input image. As illustrated in Table\ref{tab:table_us}, we can clearly see the superiority of our uncertainty strategy. We believe that other uncertainty strategies do not take into account the unique characteristics of the distribution of ground objects in RS images (dense, small targets) and appear to be inapplicable. Relatively speaking, we believe that the uncertainty in RS images is usually caused by insufficient understanding of hard-to-segment buildings with less frequency in the process of feature interaction, and our uncertainty-aware strategy can solve such a problem perfectly.



\begin{table}[]
\normalsize
\setlength\tabcolsep{3pt}
\caption{Ablation results about different uncertainty strategies on the test dataset.}
\label{tab:table_us}
\center
\begin{tabular}{c|cccc}
\hline
\multirow{2}{*}{Method} &\multicolumn{4}{c}{Inira (\%)}                                             
 \\ \cline{2-5} 
 & $IoU\uparrow$ & $F1\uparrow$ & $Pre\uparrow$ & $Recall\uparrow$ \\ \hline
GAN&
79.14	&88.36	&88.82	&87.60\\ 
CAVE &
\underline{80.55}	&\underline{89.09}	&\underline{89.62}	&\underline{88.56}\\ 
Ours &
\color{red}\textbf{83.08}  &\color{red}\textbf{90.76} &\color{red}\textbf{92.04} &\color{red}\textbf{89.52}
\\ \hline
\end{tabular}
\end{table}

\subsection{Complexity of UANet}

\begin{figure}
\centering
\includegraphics[width=0.9\linewidth]{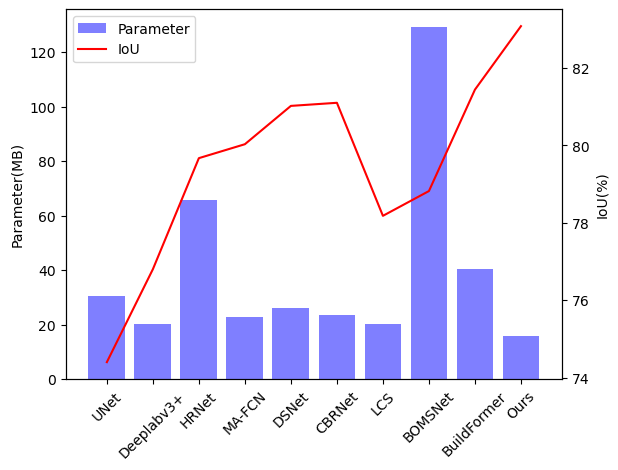}%
\caption{Complexity and accuracy comparison of our UANet and the comparison methods.}
\label{fig_parameter}
\vspace{-1.0em}
\end{figure}

In order to validate the efficiency of the proposed UANet, we compared the amount of the parameters and the IoU on the Inria building dataset with the current SOTA methods. As shown in Fig.\ref{fig_parameter}, our UANet achieves the highest accuracy with the total parameter of 15.6 M, which is the lowest.

\section{Conclusion}

In this paper, we argue that the complex distribution of the ground objects, inconsistent building scales, and various building styles bring some uncertainty to the predictions of the general deep learning models, causing the omission and the commission to a large extent. Therefore, we introduce the concept of uncertainty and propose a novel uncertainty-aware network (UANet). Firstly, we utilize a general encoder-decoder network to yield a general uncertain extraction map. Secondly, we propose the PIGM to enhance the highest-level features. Subsequently, the UAFM is proposed with the uncertainty rank algorithm (URA) to eliminate the uncertainty of features from high level to low level. Finally, the proposed UANet outputs the final extraction map with lower uncertainty. By conducting sufficient experiments, we validate the effectiveness of our UANet. The final high accuracy on three public datasets indicates that the introduction of the uncertainty concept in buildings has been extremely successful. However, although using such a way of ranking the level of uncertainty can help us get a better extraction result, how to allocate the weight adaptively of different uncertainty levels for URA is still an unsolved problem in the paper, which will be a focus of our future work.


\newpage

\vfill

\end{document}